%% file: main.tex
\documentclass[acmsmall,screen]{acmart}

\AtBeginDocument{%
  }

\usepackage[ruled,vlined,linesnumbered]{algorithm2e}
\usepackage{amsmath}
\usepackage{amsfonts}
\usepackage{amssymb}
\usepackage{algorithmic}
\usepackage{graphicx}
\usepackage{textcomp}
\usepackage{xcolor}
\usepackage{xspace}
\usepackage{booktabs}
\usepackage{tcolorbox}
\usepackage{diagbox}
\usepackage{hyperref}
\usepackage{lineno}
\usepackage{multirow}
\usepackage{float}
\usepackage{multicol}
\usepackage{subcaption}
\usepackage{tcolorbox}
\usepackage{fontawesome5}
\usepackage{tikz}
\usetikzlibrary{arrows.meta, positioning, calc, fit}

\tcbset{
  prompt/.style={
    colback=gray!10,
    colframe=gray!40,
    coltitle=black,
    fonttitle=\bfseries,
    boxrule=0.3pt,
    sharp corners,
    left=2mm,
    right=2mm,
    top=1mm,
    bottom=1mm,
  }
}

\definecolor{darkgreen}{RGB}{0,100,0}

\newcommand{\tool}{\textsc{HyNeA}\xspace} %
\newcommand{\mimicry}{\textsc{Mimicry}\xspace} % 
\newcommand{\mrm}{\textsc{GIFTbench}\xspace}

\newcommand{\maybeicon}[2]{%
  \ifstrequal{#1}{icon}{#2\,}{}%
}
\newcommand{\yolo}[1][noicon]{\maybeicon{#1}{\textcolor{darkgray}{\faCar}}\textsc{Driving}\xspace}
\newcommand{\imagenet}[1][noicon]{\maybeicon{#1}{\textcolor{darkgray}{\faImage}}\textsc{ImageNet}\xspace}
\newcommand{\celeba}[1][noicon]{\maybeicon{#1}{\textcolor{darkgray}{\faUser}}\textsc{CelebA}\xspace}

\newcounter{promptctr}
\renewcommand{\thepromptctr}{P.\arabic{promptctr}}

\renewcommand*{\equationautorefname}{Equation}
\def\equationautorefname~#1\null{(#1)\null}

\newenvironment{promptbox}[1][]{
  \refstepcounter{promptctr}
  \begin{tcolorbox}[prompt, title=Prompt~\thepromptctr, #1]
}{
  \end{tcolorbox}
}

\RestyleAlgo{ruled}
\SetKwComment{Comment}{\# }{}
\newlength\mylen

\DeclareMathOperator*{\argmax}{argmax\;} 

\DeclareMathOperator*{\argsort}{argsort\;}

\newcommand{\head}[1]{\noindent\textbf{#1.}}

\newboolean{showcomments}
\setboolean{showcomments}{true}
\ifthenelse{\boolean{showcomments}}
{
	\definecolor{myyellow}{RGB}{255, 228, 26}
	\definecolor{myblue}{RGB}{50, 50, 220}
	\newcommand{\nb}[2]{
		{\sf
			\fcolorbox{myyellow}{yellow}{\scriptsize\textbf{#1}}%
			$\blacktriangleright$%
			{\color{myblue}\fontsize{7pt}{8pt}\selectfont\textbf{#2}}%
		}%
	}
}
{
	\newcommand{\nb}[2]{}
}

\begin{document}

\title{HyperNet-Adaptation for Diffusion-Based Test Case Generation}

\author{Oliver Wei{\ss}l}
\orcid{0009-0008-7575-0187}
\email{o.weissl@tum.de}
%\authornotemark[2]
\affiliation{%
  \institution{Technical University of Munich}
  \city{Garching near Munich}
  \country{Germany}
}

\author{Vincenzo Riccio}
\orcid{0000-0002-6229-8231}
\email{vincenzo.riccio@uniud.it}
%\authornotemark[3]
\affiliation{%
  \institution{University of Udine}
  \city{Udine}
  \country{Italy}
}

\author{Severin Kacianka}
\orcid{0000-0002-2546-3031}
\email{severin.kacianka@gmail.com}
%\authornotemark[2]
\affiliation{%
  \institution{Independent Researcher}
  \country{Germany}
}

\author{Andrea Stocco}
\orcid{0000-0001-8956-3894}
%\authornotemark[1]
\email{andrea.stocco@tum.de}
\affiliation{%
  \institution{Technical University of Munich}
  \city{Garching near Munich}
  \country{Germany}
}
\email{stocco@fortiss.org}
%\authornotemark[2]
\affiliation{%
  \institution{fortiss GmbH}
  \city{Munich}
  \country{Germany}
}

\renewcommand{\shortauthors}{Wei{\ss}l et al.}

\input{sec/0_abstract}  
\begin{CCSXML}
<ccs2012>
   <concept>
       <concept_id>10010147.10010257</concept_id>
       <concept_desc>Computing methodologies~Machine learning</concept_desc>
       <concept_significance>500</concept_significance>
       </concept>
   <concept>
       <concept_id>10011007.10011074</concept_id>
       <concept_desc>Software and its engineering~Software creation and management</concept_desc>
       <concept_significance>300</concept_significance>
       </concept>
 </ccs2012> 
\end{CCSXML}


\received{21 January 2026}
\received[revised]{19 September 2025}
\received[accepted]{28 September 2025}

\maketitle
  
\input{sec/1_intro}
\input{sec/2_background}
\input{sec/3_method}

\input{sec/4_study}
\input{sec/5_discussion}
\input{sec/6_related_work}
\input{sec/7_conclusion}

\bibliographystyle{ACM-Reference-Format}
\bibliography{main}
\newpage
\input{sec/8_appendix}
\end{document}

%% file: sec/0_abstract.tex
\begin{abstract}
The increasing deployment of deep learning systems requires systematic evaluation of their reliability in real-world scenarios. Traditional gradient-based adversarial attacks introduce small perturbations that rarely correspond to realistic failures and mainly assess robustness rather than functional behavior. Generative test generation methods offer an alternative but are often limited to simple datasets or constrained input domains. Although diffusion models enable high-fidelity image synthesis, their computational cost and limited controllability restrict their applicability to large-scale testing. We present \tool, a generative testing method that enables direct and efficient control over diffusion-based generation. \tool provides dataset-free controllability through hypernetworks, allowing targeted manipulation of the generative process without relying on architecture-specific conditioning mechanisms or dataset-driven adaptations such as fine-tuning. \tool employs a distinct training strategy that supports instance-level tuning to identify failure-inducing test cases without requiring datasets that explicitly contain examples of similar failures. This approach enables the targeted generation of realistic failure cases at substantially lower computational cost than search-based methods. Experimental results show that \tool improves controllability and test diversity compared to existing generative test generators and generalizes to domains where failure-labeled training data is unavailable.
\end{abstract}

%% file: sec/1_intro.tex
\section{Introduction}\label{sec:intro}

Deep learning (DL) models have become central to a wide range of vision applications, from object recognition, classification and segmentation to autonomous systems~\cite{riccio2020testing}. As these models are increasingly deployed in real-world settings, evaluating their reliability to input variations and realistic scenarios becomes critical. Traditional gradient-based adversarial attacks can reveal vulnerabilities but generate perturbations that are imperceptible and may be unrealistic, failing to anticipate the diverse set of failures that may occur during operation~\cite{buzhinsky2023metrics,2020-Humbatova-ICSE}. Generative test generation approaches have emerged as an alternative, synthesizing diverse input variations that better mimic realistic scenarios~\cite{weissl2024targeted, maryam2025benchmarking, mozumder2025rbt4dnn, kang2020sinvad}. However, existing generative test generation methods are typically limited to simple datasets~\cite{2023-Riccio-ICSE}, due to architectural limitations of earlier generative models such as GANs and VAEs, whose representational capacity is insufficient for higher fidelity data~\cite{karras2019style, sauer2022styleganXL, kingma2013auto}, or rely on curated training data~\cite{maryam2025benchmarking}, restricting their applicability to complex, high-dimensional vision tasks.

Diffusion models (DM) have recently shown strong capabilities in high-fidelity image generation and therefore offer potential for functional testing through systematic generation of test scenarios~\cite{maryam2025benchmarking, mozumder2025rbt4dnn}. Yet, their high computational cost and challenges in controlling the generation process restrict their effective use. Most current diffusion-based approaches rely on prompt perturbations or simple noise manipulations ~\cite{maryam2025benchmarking, mozumder2025rbt4dnn,missaoui2023semantic}, which provide limited control over the resulting test cases.

A representative approach for introducing fine-grained control into DMs is ControlNet~\cite{zhang2023adding}, which augments a pretrained DM with an auxiliary network that modulates its intermediate representations. By injecting external conditioning signals, ControlNet steers the generative process while keeping the base model parameters fixed, enabling controlled synthesis under various forms of guidance.
However, such approaches typically rely on a priori curated conditioning data that explicitly encode the structures or attributes to be imposed during generation. As a result, developers must provide concrete examples of the desired outputs, which is challenging for high-dimensional inputs such as images. In existing work, this guidance is specified through model-based constraints, handcrafted prompts, or learned seed representations~\cite{maryam2025benchmarking, mozumder2025rbt4dnn}. These mechanisms offer limited control over the resulting outputs and provide no guarantees that the generated samples will faithfully reflect the developer's intended properties.
This limits the suitability of such methods for DL testing, where the goal is to generate previously unseen inputs that expose unanticipated model behaviors beyond those represented in the available training data.

We introduce \tool (\textbf{Hy}per\textbf{Ne}t \textbf{A}daptation), a diffusion-based generative testing method that enables dataset-free, controllable input generation via instance-specific adaptation. The method builds on the architectural principles of ControlNet by using an auxiliary hyper-network (HyperNet) to modulate a pretrained diffusion model, but fundamentally differs in how the control signal is obtained. Instead of learning a global mapping from conditioning inputs to guided outputs, \tool optimizes the HyperNet on a per-test-case basis by back-propagating from the diffusion model's output to produce conditioning signals that exhibit a desired failure in the system under test (SUT).
This instance-wise adaptation eliminates the need for curated source–target pairs and avoids large-scale retraining, while still enabling fine-grained, objective-driven control over the generated test inputs.

We evaluate \tool across multiple image-based learning tasks and compare it against state-of-the-art generative testing baselines based on latent recombination and latent perturbation. While these approaches can induce mispredictions, they often introduce visible artifacts or structural distortions that reduce the interpretability of the resulting failures. In contrast, \tool generates visually coherent and semantically plausible test inputs that induce targeted failures without compromising input quality. Across image classification and object detection tasks, \tool identifies 20--100\% more relevant failure-inducing test cases at comparable runtime. Moreover, human evaluation rates the generated inputs as up to 90\% more realistic, and image-based quality metrics indicate 40--100\% less visual degradation compared to existing methods. These results show that instance-specific HyperNet adaptation enables effective and realistic test generation with DM.

The contributions of this paper are as follows:
\begin{itemize}
    \item \noindent \textbf{Testing Method.} We propose \tool, a novel diffusion-based generative testing approach that enables dataset-free, controllable input generation through instance-specific HyperNet adaptation, avoiding the need for curated conditioning datasets or large-scale retraining~\cite{replication-package}.
    \item \noindent \textbf{Empirical Evaluation.} We conduct an empirical study across multiple vision tasks, demonstrating that \tool generates realistic and semantically coherent test cases that reveal more informative model failures than existing generative testing baselines.
\end{itemize}

%% file: sec/2_background.tex
\section{Background}\label{sec:background}

\subsection{Generative-based Test Generation for DL Systems}
Test generation for deep learning (DL) systems has increasingly shifted toward generative approaches~\cite{maryam2025benchmarking}, which aim to synthesize new test inputs by learning an implicit model of the input distribution directly from data. Rather than operating through explicit input-space perturbations or handcrafted domain models, generative methods sample or manipulate a learned latent space,\footnote{A latent space is a learned low-dimensional representation in which complex inputs are encoded such that semantic variations correspond to structured transformations (see~\cite{kingma2013auto}).} enabling the creation of novel, in-distribution test cases that are not constrained by the availability or structure of seed inputs.

Typical generative models include Variational Autoencoders (VAEs), Generative Adversarial Networks (GANs), and diffusion models~\cite{maryam2025benchmarking}. By producing new, in-distribution inputs without relying on handcrafted models or direct input perturbations, generative approaches enable broader exploration of the input space. Advances in generative modeling, particularly in scalability and sample quality~\cite{weissl2024targeted, maryam2025benchmarking, sinvad-tosem, mozumder2025rbt4dnn, dola2024cit4dnn}, have therefore increased their relevance for testing modern DL systems. In the remainder of the section, we describe diffusion models and how controlled test generation can be achieved via the ControlNet architecture.

\subsection{Diffusion Models}\label{sec:diffusion-model}

\begin{figure}[t]
    \centering
    \includegraphics[width=0.5\linewidth]{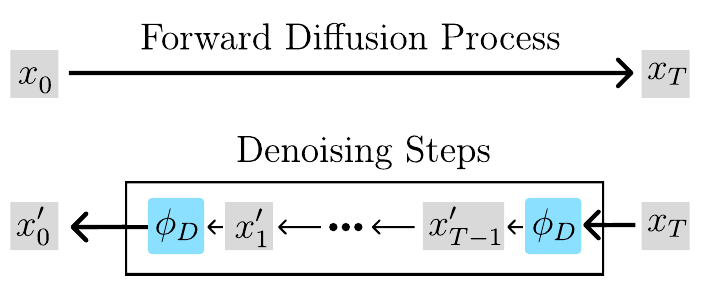}
    \caption{The diffusion and denoising process in a diffusion model.}
    \label{fig:diff}
\end{figure}

Diffusion models are a class of generative models that learn to synthesize data by reversing a gradual noise addition process. During training, data samples \(x_0\) are progressively perturbed by adding Gaussian noise over a sequence of \(T\) time steps, such that
\[
x_{n+1} = x_n + \epsilon_n, \quad \epsilon_n \sim \mathcal{N}(0, \sigma_n^2),
\]
until the final sample approximates random noise, \(x_T \approx \mathcal{N}(0, I)\) (\autoref{fig:diff}). A denoising model \(\phi_D\) is trained to predict either the added noise or the original clean data at each time step. During inference, generation starts from random noise and iteratively applies the learned denoising steps in reverse order, transforming noise into a data sample.

Originally introduced as a method for estimating data distributions~\cite{sohl2015deep}, diffusion models were later adapted to image generation tasks~\cite{song2019generative}. The introduction of a standardized training objective and explicit noise schedules~\cite{ho2020denoising,song2020denoising} significantly improved training stability and sample quality. These advances enabled diffusion models to scale from simpler benchmarks such as CIFAR-10 to higher-resolution image datasets such as ImageNet, CelebA and beyond.

Further scalability was achieved with Latent Diffusion Models (LDMs), which perform the diffusion process in a learned latent space using an encoder–decoder architecture~\cite{rombach2022high}. Operating in latent space reduces computational cost and facilitates training on high-resolution data. More recent approaches combine diffusion models with transformer-based architectures to further increase model capacity and generation quality~\cite{peebles2023scalable,leng2025repa}. Owing to their stability and expressiveness, diffusion models have also been applied to software testing in various domains~\cite{maryam2025benchmarking, mozumder2025rbt4dnn, baresi2025efficient}.

%\Vincenzo{Brief list of applications of diffusion in software testing (e.g., Davide's ICSE, Maryam's GIFTbench)}

\subsection{ControlNet}\label{sec:controlnet}

\begin{figure}[t]
    \centering
    \includegraphics[width=0.9\linewidth]{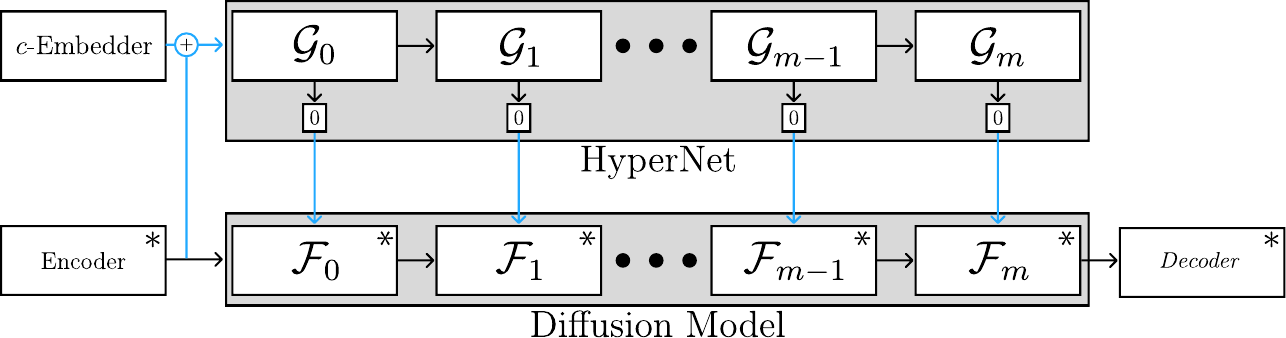}
    \caption{Simplified ControlNet Architecture (HyperNet + LDM), $\ast$ indicates Frozen Parameters, the \textcolor[HTML]{20A9FF}{blue} arrows indicate control modulation of blocks in the denoising network. For more details on architecture refer to Zhang et al.~\cite{zhang2023adding}.}
    \label{fig:controlnet}
\end{figure}

A widely adopted approach for controllable image generation in diffusion models is ControlNet~\cite{zhang2023adding}, which extends LDMs by adding an auxiliary network (\emph{HyperNet}) that modulates the generative process using a control signal $c$ (\autoref{fig:controlnet}). The control can be spatially aligned with the latent input or embedded via a conditioning function $\phi_c(\cdot)$, and is incorporated into the denoising process to guide generation toward specific structural or semantic features. 
This mechanism has proven effective for tasks such as pose-guided synthesis, edge-conditioned generation, and other fine-grained image manipulations~\cite{zhang2023adding, baresi2025efficient}, making it a natural foundation for methods aiming at controllable test-case generation in vision-related problems. 

Training a ControlNet requires a curated dataset $\mathcal{D} = \{d_i\}_{i=1}^N, \;
d_i = (X_i, y_i, c_i) \in\mathcal{X}\times\mathcal{Y}\times\mathcal{C}$, where each image $X_i$ 
is explicitly paired with a condition for generation $y_i$ (i.e a prompt), and a control signal $c_i$ encoding the structural or semantic constraints to be enforced during generation. This pairing establishes a fixed relationship between input data and the form of control available to the model.
The training objective follows the standard diffusion formulation, but with an important architectural constraint: the parameters of the pretrained diffusion backbone are kept frozen to keep existing generation capabilities of the network. The actual learning is confined to an auxiliary control branch, which adds controllability without diminishing generative expressiveness of the backbone. The optimization objective can be written as

\begin{equation}
\mathcal{L} =
\mathbb{E}_{d\in\mathcal{D},\,t,\,\epsilon\sim\mathcal{N}(0,I)}
\big[\|\epsilon - \epsilon_\theta(d,t)\|_2^2\big],
\end{equation}

where $\epsilon_\theta(\cdot)$ denotes the diffusion model augmented with the control branch.
At each denoising step, generation proceeds by combining the original diffusion transformation with an additive, control-dependent update:

\begin{align}\label{eq:controlnet}
z^{(j)} =
\mathcal{F}_j(z^{(j-1)}; y, t)
+
\gamma\,\mathcal{G}_j(z^{(j-1)}_c; y, t),
\qquad
z^{(j-1)}_c = z^{(j-1)} + \phi_c(c),
\end{align}

where $\mathcal{F}_j$ is a frozen layer of the base model and $\mathcal{G}_j$ is a layer of the ControlNet, which is influenced by the control signal. The parameter $\gamma$ regulates control strength: setting $\gamma=0$ yields unconstrained generation, while larger values increasingly bias generation toward the specified control. In practice, $\gamma$ is realised using \emph{zero layers} (indicated as $0$ in \autoref{fig:controlnet}). These zero layers are initialized with weights set to zero and are adapted during training. By updating their weights, they effectively modulate the behavior of the control network, as determined by $\gamma$ in \autoref{eq:controlnet}.  

From a testing perspective, this formulation highlights a key limitation of ControlNet-style approaches: controllability can only be learned from explicitly paired image-control examples. Consequently, the space of generable test inputs is bounded by the coverage of the conditioning data, limiting applicability when the testing objective is to explore previously unseen behaviors beyond the available training distribution. This is where \tool excels: by adapting the loss and network updates, it does not depend on curated datasets (i.e., it is \emph{dataset-free}), as described in the later sections.

%% file: sec/3_method.tex
\section{Methodology}
\label{sec:methodology}

\tool is a dataset-free method for controllable image generation that builds on the principles of ControlNet (\autoref{sec:controlnet}). While ControlNet introduces an additional network to inject control information into the diffusion process and is trained on paired data to learn mappings of the form $(c \to X)$, \tool inverts this setting. Instead of generating images from control inputs, \tool adjusts the generation process so that the produced image yields a desired control outcome $(X \to c)$, defined through user-specified optimization objectives that steer the generation process. This reversal removes the dependency on curated datasets and allows \tool to guide generation based solely on the behavior observed in the produced outputs, making it suitable for targeted test case generation for DL systems.

In contrast to augmentation-based approaches that require an existing input image and subsequently manipulate it, \tool generates images directly from the diffusion prior starting from random noise (initial latent $z$). The optimization process does not depend on pre-existing test inputs or seed images. Instead, \tool first synthesizes an image from scratch and then iteratively adapts the generation process until the produced sample induces the targeted SUT behavior.

The architecture of \tool consists of four components (see \autoref{fig:hynea_loop}):

\begin{itemize}
\item \textbf{SUT}: The system under test whose behavior is evaluated.
\item \textbf{Manipulator}: The LDM and HyperNet used to produce and manipulate test cases. %\Andrea{isn't always a DMs?}
\item \textbf{Optimizer}: The optimizer used to adapt weights in the HyperNet used in \tool during generation of test cases. %\Andrea{isn't always a DMs?}
\item \textbf{Objectives}: The loss function that defines the criteria for successful test case generation. 

\end{itemize}

\input{img/flow}
\tool replaces the supervised reconstruction objective used in ControlNet-style training with a \emph{behavior-driven feedback loop}. Rather than enforcing a fixed pairing between training images and control signals, the model is evaluated based on the properties of its final generated outputs.

Given an initial control signal $c_0 \in \mathcal C$, the Manipulator produces a final image $\hat X \sim p_\theta(\cdot \mid c_0)$. A corresponding control value $\hat c \in \mathcal C$ is extracted from $\hat X$ using the SUT's prediction. We define a collection of loss terms $\{\omega_i\}_{i=1}^K$, each evaluating properties of the generated image that are targeted in the optimization. The overall training objective is defined as:

\begin{equation}\label{eq:loss}
\mathcal L =
\mathbb E_{\hat X \sim p_\theta(\cdot \mid c_0)}
\left[
\sum_{i=1}^K \alpha_i\,\omega_i(\cdot)
\right].
\end{equation}

A key distinction from standard ControlNet-based diffusion training lies in how the loss is applied throughout the diffusion process. Conventional methods compute losses independently at each denoising step and backpropagate gradients locally, keeping memory usage low. In contrast, \tool evaluates the alignment objective with respect to the final generated image, aggregating the influence of the control signal across all denoising steps. This yields more informative gradients for guiding generation but increases memory demand, making techniques such as gradient checkpointing necessary to keep training feasible.

When multiple loss terms are used, the coefficients $\alpha_i$ allow for fine-grained control over their relative contributions. In this work, we set $\alpha_i = 1$ for all $i$, such that no explicit reweighting is applied and each loss term contributes equally to the overall objective.

\subsection{System under Test}

We consider three vision tasks: multiclass classification, binary classification, and object detection. In all cases, the SUT $\phi_P(\cdot)$ maps an input image $X$ to a prediction $y$, 
\[
X \xrightarrow{\phi_P} y.
\]
The objective of \tool is to generate realistic, failure-inducing test cases \(X'\) that preserve the semantic ground-truth label but cause the SUT to predict a different label:
\[
\mathrm{oracle}(X') \neq \phi_P(X')
\]

\head{Multiclass Classification} For standard classification, the output is $y \in \mathbb{R}^C$, where $C$ is the number of classes. The predicted class is obtained by $\hat{c} = \arg\max y$.

\head{Binary Classification} For binary attribute prediction, the output is either a single logit $y \in \mathbb{R}$ (one attribute) or a vector $y \in \mathbb{R}^C$ (multiple independent attributes). Unlike multiclass classification, attributes are evaluated independently: an attribute $c$ is predicted as present if $y_c > 0$.

\head{Object Detection} We focus on one-stage dense object detectors, which predict a large number of candidate objects per image.
Accordingly, the model outputs $y \in \mathbb{R}^{D \times C}$, where $D$ is the number of detections and $C$ the number of classes. Each detection $d$ has an associated class distribution; its predicted class is $\hat{c}_d = \arg\max y_d$.
Detections can be ranked by their maximum confidence values, producing an ordered list from the most to the least confident predictions. We follow this ranking in our implementation and restrict evaluation to the top-5 detections. This restriction is necessary because object detectors can produce thousands of detections per image~\cite{Jocher_Ultralytics_YOLO_2023}, while our optimization relies on a single scalar loss value. Aggregating a large number of detections into one loss causes the contribution of individual detections to be heavily diluted, leading to vanishing controllability during optimization.

\subsection{Manipulator}

The manipulators in \tool are LDMs with varying architectures. While the specific architecture is not critical, it is important that we can copy the weights of relevant blocks to initialize the HyperNet. The original networks are frozen and do not change, but the copied HyperNet is trainable.  

To manipulate the latent states of the LDM, we add \emph{zero layers}, which are inserted between a copied block and a standard block to modulate the copied network's outputs (see \autoref{sec:controlnet}).

Unlike standard ControlNets, which require control signals to be spatially aligned with the input image, \tool directly uses the SUT's predictions as control inputs. Since these predictions may be non-spatial, we introduce a trainable condition embedding function $\phi_c(\cdot)$ that projects arbitrary SUT outputs into a spatial control signal compatible with the image latent space. The projection mechanism and its dimensionality-dependent cases are formalized in \autoref{alg:proj} and further described in \autoref{sec:projector}.

\subsubsection{Control Projector}\label{sec:projector}
The SUT's output $\hat{y}$ may vary significantly in structure depending on the task, such as sequences of vectors (e.g., object detection scores), or flat vectors (e.g., classification outputs). To use these signals as conditioning inputs for \tool, they must be transformed into a tensor with the same spatial dimensionality as the image latent $X$.

The control projector $\phi_c$ therefore performs a dimensionality-aware projection. If $\hat{y}$ already matches the spatial dimensionality of $X$, a convolution is used to adapt channel dimensions while preserving spatial structure. If $\hat{y}$ is missing exactly one spatial dimension, convolution is applied along the existing dimensions, followed by a linear expansion of the missing dimension. For fully non-spatial outputs, a linear projection is used to expand the signal into a spatial representation. In all cases, the resulting tensor is reshaped to match the spatial dimensions of $X$, yielding a control signal that can be injected into the diffusion process.

\begin{algorithm}[t]
\footnotesize
\caption{Control Projector $\phi_c$}\label{alg:proj}
\KwIn{$\hat{y}$ \Comment*[r]{SUT's prediction}}
\KwOut{$c$ \Comment*[r]{Control signal projected to match spatial dimensions of $X$}}

\If{$\hat{y}.\operatorname{ndim} = X.\operatorname{ndim}$}{
    $c \gets \operatorname{Conv2D}(\hat{y})$ \Comment*[r]{Convolute to match spatial size}
}
\ElseIf{$X.\operatorname{ndim} - \hat{y}.\operatorname{ndim} = 1$}{
    $c \gets \operatorname{Conv1D}(\hat{y})$ \Comment*[r]{Convolute along existing dimensions}
    $c \gets \operatorname{Linear}(c)$ \Comment*[r]{Expand missing dimension with linear layer}
}
\Else{
    $c \gets \operatorname{Linear}(\hat{y})$ \Comment*[r]{No matching dimensions; expand fully with linear layer}
}
$c \gets \operatorname{flatten}(c)$ \Comment*[r]{Flatten to a single vector}
$c \gets \operatorname{reshape}(c, X.\operatorname{shape})$ \Comment*[r]{Reshape to match $X$}
\Return{$c$ \Comment*[r]{Projected control signal aligned with $X$}}
\end{algorithm}

\subsection{Optimizer}\label{sec:optimizer}
As \tool leverages the weights of a HyperNet to generate test cases, these weights are optimized as in standard neural network training, but independently for each instance.

To optimize the weights $\theta$ in \tool, we use AdamW~\cite{loshchilov2017fixing} as the primary optimizer. AdamW has shown strong convergence properties across a variety of tasks, including classification and generative model training~\cite{leng2025repa, zhang2023adding}.  

We pair AdamW with the OneCycle learning rate schedule~\cite{smith2019super}, which accelerates convergence by varying the learning rate from a minimal value $lr_{min}$ to a maximal value $lr_{max}$ following a curve resembling a positively skewed Gaussian. At the start of training, the learning rate increases rapidly to allow fast exploration, then gradually decreases to stabilize convergence. After each backpropagation step, the scheduler updates the learning rate slightly, enabling smooth adaptation throughout training.

The values of $lr_{min}$ and $lr_{max}$ depend on the capabilities of the generative network (its ability to separate features in images) and the robustness of the SUT (how easily its predictions can be changed). Therefore, there is no universal choice; the learning rates must be selected based on the specific use case. Generally, one should consider a tradeoff between generation quality and runtime to select an appropriate schedule.

Similarly, the number of steps in the schedule varies by task, but for our experiments we use 2,500 steps, aiming to produce small, incremental changes that effectively influence SUT behavior without destabilizing the generator.

\subsection{Objectives}\label{sec:objectives}
In \tool, objectives are functions that determine the overall loss during a forward pass through the network. These objectives guide the generator toward producing test cases that meaningfully alter the SUT's behavior.

\head{Visual Fidelity} For all tasks, we include an image-based loss that quantifies how much the generated image differs from the original. We use the Frobenius distance between two images $A$ and $B$, as defined in \autoref{eq:fd}. The purpose of this loss is to constrain low-level image-space deviations between the generated image and the original input. The Frobenius distance provides a simple and computationally efficient measure of overall pixel-level change magnitude during optimization.

\begin{equation}\label{eq:fd}
    \omega_{dF}(A, B) := \sqrt{\operatorname{tr}((A-B)^\intercal(A-B))}.
\end{equation}

\head{Behavior Steering} In addition to image similarity, \tool considers the outputs of the SUT as control signals. To generate targeted failures, we enforce a loss that moves the SUT's outputs toward a desired target, representing a specific failure or behavior we want to explore.
 
For classification tasks, we use the cross-entropy loss to measure the difference between the SUT's output $\hat{y}$ and the target class $t$, as defined in \autoref{eq:cel}.

\begin{equation}\label{eq:cel}
    \omega_{CE}(\hat{y}, t) := -\hat{y}_t + \log \sum_{c=1}^{C} \exp(\hat{y}_c).
\end{equation}

where $C$ is the total number of classes. This loss encourages the SUT to predict the target class while penalizing deviations.

For binary classification, we adapt the objective to binary cross-entropy (BCE) loss with logits. Here, $\hat{y}$ can be a single value if predicting a single binary class, or a vector of length $C$ for multiple binary classes. The target vector is defined as $t = C \times \{0,1\}$, indicating which classes should be positive. The BCE loss for a specific class $c$ is given in \autoref{eq:bce}.

\begin{equation}\label{eq:bce}
    \omega_{BCE}(\hat{y}_c, t_c) := - \left[ t_c \log \sigma(\hat{y}_c) + (1-t_c) \log (1 - \sigma(\hat{y}_c)) \right].
\end{equation}

where $\sigma(\cdot)$ is the sigmoid function. This formulation directly measures how far the predicted probability is from the desired binary target.

In object detection, each image can produce multiple predictions, resulting in a $D \times C$ tensor $\hat{y}$, where $D$ is the number of detected objects. We compute the mean-reduced cross-entropy loss over all detections, as shown in \autoref{eq:celm}.

\begin{equation}\label{eq:celm}
    \omega_{CEm}(\hat{y}, t) := \frac{1}{D} \sum_{d=1}^{D} \omega_{CE}(\hat{y}_d, t).
\end{equation}

This ensures that the loss considers every detection in the image, averaging their contributions to guide the generator toward producing failures across multiple objects.

In \tool, all loss terms are backpropagated through the network. For each task the \emph{Visual Fidelity} objective is combined with an appropriate \emph{Behavior Steering} objective depending on the type of SUT \autoref{eq:loss}:

\begin{itemize}
    \item \textit{Classification:} $\mathcal{L}_C = \omega_{dF} + \omega_{CE}$  
    \item \textit{Binary Classification:} $\mathcal{L}_B = \omega_{dF} + \omega_{BCE}$ 
    \item \textit{Object Detection:} $\mathcal{L}_O = \omega_{dF} + \omega_{CEm}$
\end{itemize}

This combined objective ensures that the generated images both meaningfully alter the SUT's output and remain visually coherent relative to the original image.

%% file: img/flow.tex
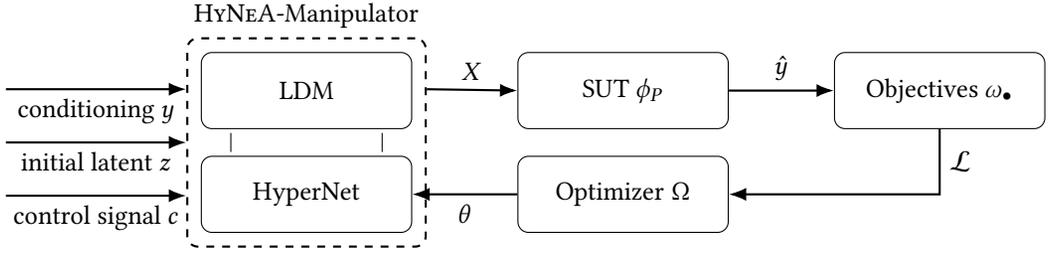
\begin{figure}[t]
    \centering
    \begin{tikzpicture}[
        block/.style={rectangle, draw, rounded corners, minimum width=2.8cm, minimum height=1.0cm, align=center},
        arrow/.style={-Latex, thick},
        node distance=1.6cm and 1.4cm
    ]

% --- HyNeA as two-row module (top row: LDM -> SUT, bottom row: ControlNet <- Optimizer) ---
\node[block] (ldm) {LDM};
\node[block, below=0.35cm of ldm] (cnet) {HyperNet};

% dashed container
\node[draw, dashed, thick, rounded corners, fit=(ldm)(cnet),
      label={\tool-Manipulator}, inner sep=0.18cm] (hyneabox) {};

% right side: align SUT with LDM (top row)
\node[block, right=of ldm] (sut) {SUT $\phi_P$};
\node[block, right=of sut] (objectives) {Objectives $\omega_\bullet$};

% bottom row: align Optimizer with ControlNet
\node[block, right=of cnet] (opt) {Optimizer $\Omega$};

% --- Inputs (left) ---
\draw[arrow] ($(hyneabox.west)+(-2.4,0.7)$) -- ($(hyneabox.west)+(0.0,0.7)$)
    node[pos=0.5, below] {conditioning $y$};

\draw[arrow] ($(hyneabox.west)+(-2.4,0.0)$) -- (hyneabox.west)
    node[pos=0.5, below] {initial latent $z$};

\draw[arrow] ($(hyneabox.west)+(-2.4,-0.7)$) -- ($(hyneabox.west)+(0.0,-0.7)$)
    node[pos=0.5, below] {control signal $c$};

% --- Indicate LDM and ControlNet work together (two internal connections) ---
\draw ($(cnet.north east)+(-0.4,0.05)$) -- ($(ldm.south east)+(-0.4,-0.05)$);
\draw ($(cnet.north west)+(0.4,0.05)$) -- ($(ldm.south west)+(0.4,-0.05)$);

% --- Forward path (top row) ---
\draw[arrow] ($(hyneabox.east)+(0.0,0.7)$) -- (sut.west) node[pos=0.5, above] {$X$};
\draw[arrow] (sut.east) -- (objectives.west) node[pos=0.5, above] {$\hat{y}$};

% --- Objectives into Optimizer (down to bottom row) ---
\draw[arrow] (objectives.south) |- (opt.east)
    node[pos=0.25, right] {$\mathcal{L}$};

% --- Parameter update (bottom row) ---
\draw[arrow] (opt.west) -- (cnet.east) node[pos=0.5, below] {$\theta$};

\end{tikzpicture}
    \caption{\tool's component interaction throughout Optimization. Each component and its respective In- and Outputs are described in the following sections.}
    \label{fig:hynea_loop}
\end{figure}

%% file: sec/4_study.tex
\section{Empirical Study}\label{sec:empirical-study}

\subsection{Research Questions}\label{sec:rqs}
\textbf{RQ\textsubscript{1} (effectiveness):} \textit{How effective is \tool in generating diverse, yet semantically faithful failure-inducing test cases?}

We investigate the effectiveness of \tool at exposing failures compared to existing baseline approaches. Beyond raw failure discovery, effective software testing requires test cases that induce controlled, meaningful changes while maintaining diversity to avoid redundant coverage.
Accordingly, we assess additional quality measures, including changes in image space and the diversity of the generated test cases from each method in the embedding space.

\noindent
\textbf{RQ\textsubscript{2} (efficiency):} \textit{How efficient is \tool in generating failure-inducing test cases?}

We assess whether \tool's efficiency in terms of budget use and runtime is competitive with respect to baseline methods. 

\noindent
\textbf{RQ\textsubscript{3} (validity):} \textit{Are generated test cases by \tool semantically valid?}

Generating test cases is only useful when they remain semantically valid, therefore we compare \tool with our baseline propositions in terms of realism and semantic preservation.

\noindent
\textbf{RQ\textsubscript{4} (sensitivity):} \textit{How sensitive is \tool's performance to hyperparameter and random seed selection?}

As \tool's methodology is based on adapting network weights, the learning rate constitutes its most influential hyperparameter. In addition, the stochastic nature of diffusion-based generation may introduce variability across random seeds. Therefore, we investigate how learning-rate and seed selection affect \tool's effectiveness, efficiency, and validity.

\subsection{Metrics \& Measures}\label{sec:metrics}

\subsubsection{Metrics used in RQ\textsubscript{1}}
\paragraph{\textbf{MS-SSIM} ($\uparrow$)}  
To quantify structural changes between the original and generated images, we use the multi-scale Structural Similarity Index (MS-SSIM)~\cite{wang2003multiscale}, an extension of SSIM that evaluates image similarity across multiple spatial scales. MS-SSIM addresses a known limitation of SSIM—its dependence on image scale and viewing distance—by repeatedly downsampling the image pair and aggregating luminance, contrast, and structural comparisons at each scale. This yields a metric that aligned more closely with human perception and is less sensitive to minor color or brightness variations. Higher MS-SSIM values indicate that the generated image remains structurally close to the original, meaning fewer structural modifications were required to produce a valid test case.

\paragraph{\textbf{Embedding Diversity} ($\uparrow$)} We also evaluate the diversity of generated images, as test case generation benefits from covering a wide range of possibilities. Directly measuring diversity in image space is challenging, so we use embeddings from a pretrained ResNet ($\phi_E(\cdot))$ on ImageNet. These embedding are produced by passing an image $X$ through the embedding network $E=\phi_E(X)$. Embeddings for similar content are close, while embeddings for more diverse content diverge. We quantify diversity as the mean variance across embeddings $E$ of multiple outputs, as shown in \autoref{eq:div}

\begin{equation}\label{eq:div}
    \operatorname{diversity}(E) := \frac{1}{|E|} \sum_{i=1}^{|E|} \operatorname{Var}(E_i).
\end{equation}

Higher values indicate that the method produces more diverse test cases.
In this context, diversity refers to generated test cases being visually distinct from one another across multiple generations.

\paragraph{\textbf{Trace Difference} ($\downarrow$)} 
While diversity should be high across samples, test cases should remain semantically close to the original inputs. To assess semantic changes, the trace difference metric compares embeddings of the original image $E_0$ and the generated test case $E_t$ using differences in the covariance of their embeddings, as defined in \autoref{eq:trace}. Intuitively, this metric captures changes in per-dimension variance while discarding cross-dimensional correlations, which is shown when reformulating it to a $\ell_1$-distance based metric as shown in \autoref{app:l1trace}.

\begin{equation}\label{eq:trace}
    \operatorname{trace\Delta}(E_0, E_t) := \operatorname{tr}(|\operatorname{cov}(E_0^\intercal)-\operatorname{cov}(E_t^\intercal)|)
\end{equation}

Here, $\operatorname{cov}(\cdot)$ denotes the covariance matrix. Lower values indicate that the semantic content of the network's input remains similar, even though a test case has been generated.

\paragraph{\textbf{Misclassification Rate} ($\uparrow$)} To assess task performance, we measure how many generated test cases change the predictions of the SUT. The same overall principle applies to classification, binary classification, and object detection, but the exact computation differs by task.
\begin{equation}\label{eq:mrate}
    \operatorname{mr}_C(\mathcal{Y}, \hat{\mathcal{Y}}) := \frac{1}{B}\sum_{i=1}^{B}
    \left[\argmax \mathcal{Y}_i \neq \argmax \hat{\mathcal{Y}}_i \right].
\end{equation}

For binary classification, we only evaluate whether the targeted class logit flips sign, indicating that the model's decision boundary has been crossed. Let \(c\) be the targeted class. The misclassification rate is given by \autoref{eq:mrate_b}.
\begin{equation}\label{eq:mrate_b}
    \operatorname{mr}_B(\mathcal{Y}_c, \hat{\mathcal{Y}}_c)
    := \frac{1}{B}\sum_{i=1}^{B}
    \left[(\mathcal{Y}_{i,c} > 0) \neq (\hat{\mathcal{Y}}_{i,c} > 0)\right].
\end{equation}

For multi-class classification, the misclassification rate is defined in \autoref{eq:mrate}. Let \(\mathcal{Y}\) denote the predictions on the initial images and \(\hat{\mathcal{Y}}\) the predictions on the corresponding test cases. Both prediction sets are tensors of shape \(B \times C\), with \(B\) denoting the number of SUT outputs and \(C\) the number of classes. The Iverson bracket \([\cdot]\) evaluates to \(1\) if its statement is true and \(0\) otherwise.

For object detection, each SUT output contains multiple detections, so predictions are of shape \(B \times D \times C\), where \(D\) is the number of detections. We compute the misclassification rate by averaging over detections as well, as shown in \autoref{eq:mrate_o}.
\begin{equation}\label{eq:mrate_o}
    \operatorname{mr}_O(\mathcal{Y}, \hat{\mathcal{Y}}) 
    := \frac{1}{BD}\sum_{i=1}^{B}\sum_{j=1}^{D}
    \left[\argmax \mathcal{Y}_{i,j} \neq \argmax \hat{\mathcal{Y}}_{i,j} \right].
\end{equation}

A higher misclassification rate indicates that more test cases induced meaningful behavioral changes in the SUT.

\paragraph{\textbf{Escape Ratio} ($\downarrow$)} Following \mimicry~\cite{weissl2024targeted}, we measure whether generated test cases respect their intended target, providing a proxy for controllability of the generation process. For classification, the escape ratio remains identical to the original formulation and is shown in \autoref{eq:escape}. Let \autoref{eq:pi} return the class ordering for the targeted class \(c\). The two highest-scoring classes in the original prediction are \(\mathcal{Y}_{c,\pi 1}\) and \(\mathcal{Y}_{c,\pi 2}\).
\begin{align}
    \operatorname{esc}_C(\mathcal{Y}, \hat{\mathcal{Y}}) 
    &:= \frac{1}{B}\sum_{i=1}^{B}
    \left[
        \argmax \hat{\mathcal{Y}}_i 
        \notin \{\mathcal{Y}_{c,\pi 1},\, \mathcal{Y}_{c,\pi 2}\}
    \right],\label{eq:escape}\\
    \pi &= \argsort \mathcal{Y}_c.\label{eq:pi}
\end{align}

For binary classification, we aim to determine whether modifying the target logit unintentionally alters other binary attributes. We therefore adapt the escape ratio to count how many non-target classes change sign, using the binary misclassification rate defined earlier. This is shown in \autoref{eq:escrb}.
\begin{equation}\label{eq:escrb}
    \operatorname{esc}_B(\mathcal{Y}, \hat{\mathcal{Y}})
    := \frac{1}{C-1}\sum_{j=1}^{C}
        \operatorname{mr}_B(\mathcal{Y}_j, \hat{\mathcal{Y}}_j)\,
        [j \neq c].
\end{equation}

For escape ratio, lower is better: a small value indicates precise control, meaning the targeted failure mode is influenced without unintentionally affecting unrelated features. A high escape ratio signals that test case generation affects additional outputs, suggesting insufficient fine-grained control.

\paragraph{\textbf{Confidence Reduction} ($\uparrow$).}
For object detection, each input produces multiple detections. To evaluate whether the generated test cases consistently reduce confidence in the original class across all detections, we measure the average decrease in the class-\(c\) logit. The confidence reduction is defined in \autoref{eq:cr}. Predictions have shape \(B \times D \times C\), with \(B\) the number of inputs and \(D\) the number of detections per input.
\begin{equation}\label{eq:cr}
    \operatorname{cr}(\mathcal{Y}_c, \hat{\mathcal{Y}}_c)
    := \frac{1}{BD}\sum_{i=1}^{B}\sum_{j=1}^{D}
        \left(\mathcal{Y}_{i,j,c} - \hat{\mathcal{Y}}_{i,j,c}\right).
\end{equation}
Higher values indicate stronger confidence reduction in the original class across all detections, showing that the test cases uniformly suppress the model's confidence in the targeted class.

\subsubsection{Measures used in RQ\textsubscript{2}}
\paragraph{\textbf{Average runtime per generated test case} ($\downarrow$).}
To assess \tool's efficiency relative to the baseline methods \mimicry and \mrm, we compare the average runtime in seconds required to generate a single test case.

\paragraph{\textbf{Computational budget} ($\downarrow$).}
We define the computational budget as the number of SUT evaluations, which makes results comparable across methods. In addition to the implemented baselines, we also extrapolate the expected runtime of a hybrid approach that combines search-based strategies like \mimicry with diffusion models.

\paragraph{\textbf{TFLOPs per generated test case} ($\downarrow$).}
To provide a hardware- and architecture-agnostic estimate of computational cost, we additionally report the average tera floating point operations (TFLOPs) required to generate a single test case. This complements the runtime-based analysis by quantifying the underlying computational workload independently of implementation or hardware differences.

\subsubsection{Measures used in RQ\textsubscript{3}}

\paragraph{\textbf{Label preservation} ($\uparrow$).}
Label preservation is defined as the fraction of generated test cases that human evaluators judge to retain the original class label. A higher score indicates that a larger proportion of the generated test cases remain semantically valid to human annotators, suggesting that any observed change in the SUT's behavior corresponds to a genuine failure rather than a semantic shift in the input.

\paragraph{\textbf{Image realism} ($\uparrow$).}
Image realism captures the overall perceived realism of generated images as assessed by human evaluators on a normalized scale from 0 (very unrealistic) to 1 (perfect realism). This metric quantifies the visual quality of the generated images: while object presence may still be identifiable in stylized or cartoon-like images, such cases are arguably not valid test inputs for the SUT.

\paragraph{\textbf{Ambiguity score} ($\downarrow$).}
The ambiguity score is defined as the fraction of generated images for which human evaluators could not decisively determine whether the original class label was preserved (i.e., marked as \emph{Unsure}). Lower ambiguity is preferred, as a high ambiguity score may indicate that generated test cases fall outside the valid input domain and are nonsensical even to human observers.

\paragraph{\textbf{Inter-rater agreement} ($\uparrow$).}
We report Fleiss' $\kappa$ to quantify inter-rater agreement among human evaluators in the annotation study. Higher values indicate stronger agreement between annotators, whereas low agreement may suggest poor response quality or the presence of test inputs that are difficult or impossible for humans to interpret reliably.

\subsubsection{Metrics \& Measures used in RQ\textsubscript{4}}
RQ\textsubscript{4} uses a combined set of metrics from RQ\textsubscript{1}-RQ\textsubscript{3} (MS-SSIM, image realism, and runtime). These metrics are evaluated jointly to identify the best trade-off across different learning-rate configurations. In addition, we include the FID score to enable trade-off assessment without relying on human annotators.

To assess sensitivity to random seed selection, we analyze the distributions of MS-SSIM, LPIPS, and runtime across multiple seed configurations using kernel density estimation (KDE). Furthermore, we apply the Anderson--Darling k-sample test~\cite{scholz1987k} to determine whether the observed metric distributions differ significantly across seeds. Together, these analyses provide insight into the robustness of \tool with respect to both hyperparameter and seed selection.

\subsection{Baseline Methods}

We compare our approach against two recent generative–based test case generation methods, \mimicry~\cite{weissl2024targeted} and \mrm~\cite{maryam2025benchmarking}.

\mimicry is a StyleGAN-based method that generates targeted test cases by mixing latent vectors within the StyleGAN latent space, enabling controlled generation toward decision boundaries of the SUT. It first samples an initial latent vector and renders the corresponding image. Based on the SUT's second most likely class for this image, \mimicry then samples a second latent vector associated with the target class. These two vectors are interpolated using a population of interpolation vectors that are optimized toward via a multi-objective optimizer. In our work, to guarantee a fair comparison, we exchange \mimicry's objective functions with the loss terms defined in \autoref{sec:objectives}. %\Vincenzo{''to guarantee a fair comparison''?} 

\mrm, in contrast, is a diffusion-based method for generating failure cases by manipulating the latent representation of a diffusion model within a population-based optimization framework. It employs single-point crossover to recombine latent vectors and mutation operators that inject noise at varying scales.
The method relies on fine-tuned instances of Stable Diffusion~1.5, which are guided by natural-language prompts to generate the corresponding outputs. While the original work did not consider object detection as a testing domain~\cite{maryam2025benchmarking}, the approach can be readily extended by adapting the fitness calculation, while all other components remain unchanged.

\subsection{Objects of Study}
We evaluate our approach across three popular DL vision tasks, namely classification, binary classification, and object detection, each paired with a task-specific diffusion generator and a corresponding SUT.
\subsubsection{Multi-Class Classification}%\Andrea{consider renaming it to Multi-class Classification}

We adopt the ImageNet-1k dataset, a standard benchmark for large-scale image classification which contains 14 million images across 1,000 classes. Its diversity and visual complexity motivate the use of diffusion models that can capture fine-grained visual structure. We focus on ten classes: \textit{goldfish, great white shark, tiger shark, hammerhead shark, electric ray, stingray, cock, hen, teddy bear, pizza}. These overlap with the classes used in \mimicry, whereas \mrm supports two of them (\textit{teddy bear}, \textit{pizza}).

\head{SUT}
We use pretrained WideResNet-50~\cite{zagoruyko2016wide}, EfficientNetV2-M~\cite{tan2021efficientnetv2}, and ViT-L-16~\cite{dosovitskiy2020image} models from the \texttt{torchvision} library \cite{torchvision2016} to ensure reproducibility and a fair comparison, as standardized pretrained models are also used in the evaluation of \mimicry and \mrm.

\head{Diffusion Model}
We employ a transformer-based LDM of the REPA-E family, which has demonstrated state-of-the-art generative performance on ImageNet benchmarks~\cite{leng2025repa}. REPA-E extends the REPA training paradigm, by including the encoder and decoder networks in training, which makes it well suited for high-fidelity, class-conditional generation.

\subsubsection{Multi-label Binary Classification}

CelebA provides facial images annotated with 40 binary attributes, making it a standard benchmark for multi-attribute prediction. We select ten attributes for detailed evaluation: \textit{Arched Eyebrows, Big Lips, Big Nose, Chubby, Eyeglasses, Goatee, Heavy Makeup, Male, Smiling, Wearing Hat}. This selection is intended to cover a range of attributes at varying levels of granularity and overlaps with existing work by \mrm~\cite{maryam2025benchmarking}.

\head{SUT}
We use a ResNet50-based attribute classifier pretrained on ImageNet via \texttt{torchvision}~\cite{torchvision2016}. The model is fine-tuned on the CelebA training split and outputs 40 logits, one per attribute, effectively functioning as an ensemble of binary classifiers~\cite{2026-Chen-DETECT}.

\head{Diffusion Model}
We use a pretrained UNet-based LDM~\cite{ldmcelebahq2022}, following the Stable Diffusion architecture~\cite{rombach2022high}. We select this network because it is pretrained and publicly available, which ensures good reproducibility and avoids introducing training bias that could arise from tuning a custom generative model.
Although this architecture is less recent and has slightly lower capacity than transformer-based models such as REPA-E models~\cite{leng2025repa}, it nevertheless demonstrates that our method applies consistently across different families of generative models.

\subsubsection{Object Detection}
We consider object detection in a driving context, focusing on urban road scenes with multiple interacting objects. The benchmark setting is based on real-world driving imagery and semantic segmentation annotations derived from the KITTI dataset~\cite{Geiger2013IJRR} and a custom dataset located in Munich, Germany~\cite{cccccsys_guericke_dataset_2}. This task reflects the safety-critical nature of perception systems used in autonomous driving and enables evaluation on complex, structured scenes.

\head{SUT}
We use YOLOv8~\cite{Jocher_Ultralytics_YOLO_2023}, chosen due to its strong accuracy–speed tradeoff and its widespread use in autonomous driving research~\cite{tian2024pedestrian, wibowo2023object, yu2022traffic, jiang2025fm, im2022adversarial, lambertenghi2024assessing}.

\head{Diffusion Model}
For test generation, we rely on a Stable Diffusion 1.5 model fine-tuned on real-world road scenes~\cite{cccccsys_guericke_dataset_2}, implemented via the \texttt{diffusers} library~\cite{von_Platen_Diffusers_State-of-the-art_diffusion}. Generation is guided using a segmentation-based ControlNet~\cite{zhang2023adding} trained on driving-scene segmentations, with KITTI-derived segmentation maps serving as control signals. This setup reflects a standard diffusion pipeline used in practice and allows structured control over generated driving scenarios.

\subsection{Experimental Setup}

All experiments use pre-trained diffusion models across \imagenet[icon]~\cite{leng2025repa}, \celeba[icon]~\cite{rombach2022high}, and \yolo[icon]~\cite{cccccsys_guericke_dataset_2}.
The benchmark datasets are used to define the evaluation domains, labels, and SUT configurations; the test cases are not taken from these datasets, but are synthesized by the generative pipeline.
We begin by generating an initial origin image: \imagenet samples are conditioned on a class index. \celeba samples use no conditioning, since facial attributes can be evaluated on the generated image itself and then inverted. \yolo samples use a positive prompt (\autoref{prompt:posp}) and a negative prompt (\autoref{prompt:negp}).
For \imagenet and \yolo, we verify that the initial generation is valid by checking whether the conditioned class is still detected by the SUT; for \yolo, the class must appear as the most likely class in at least one of the top-5 detections.
For the \mimicry experiments, we use a population size of 100 individuals and optimize for 25 generations, resulting in a theoretical maximum of 2{,}700 evaluations when accounting for separate evaluations of the initial and final populations. We select this budget based on \mimicry's convergence study, which indicates that the most significant decrease in loss occurs within the first 20 generations (corresponding to a budget of 2{,}100 evaluations), suggesting that extending optimization to 25 generations should yield stable and high-quality results.
Competing methods are constrained to the same computational budget. For \mrm we extend the tasks to the driving dataset by using the same pretrained model as in \tool, with the same prompts (\autoref{prompt:posp}, \autoref{prompt:negp}).

\textbf{For RQ\textsubscript{1} and RQ\textsubscript{2}}, we collect 10 test cases per class per method. Both \imagenet and \celeba each contain 10 classes, whereas \yolo contains 5 classes. For \imagenet and \celeba, we evaluate the three methods \tool, \mimicry, and \mrm, resulting in \(3 \times 10 \times 10 = 300\) test cases per dataset. For \yolo, we evaluate only \tool and \mrm, as \mimicry does not provide a suitable generator for this domain, resulting in \(2\times 5 \times 10 = 100\) \yolo test cases. Overall, this results in a total of 700 test cases. All methods incorporate early termination conditions based on misbehavior of the SUT relative to the original output (misclassification). For each test case, we collect runtime, original images, and their corresponding targets.

\textbf{For RQ\textsubscript{3}}, we assess the realism of generated test cases and whether they preserve their original labels. We use human annotators on \imagenet classes only, as this dataset provides sufficient volume for statistical significance while keeping the study within budget. Annotators are compensated at approximately $2\$$ per completed survey, which supports response quality.
To reduce the risk of unrepresentative examples, we construct two separate batches of images and assign annotators to one of the two batches at random. Each batch contains one image per class–method pair, yielding 23 questions: 10 classes for \tool, 10 for \mimicry, and 2 for \mrm (pizza and teddy), plus 1 attention-check question. For each image, annotators rate visual realism on a 1–5 scale and indicate whether an instance of class X is present (Yes/No/Unsure). We compare methods based on perceived realism and label preservation.

\textbf{For RQ\textsubscript{4}}, motivated by the sensitivity of gradient-based adaptation in diffusion models discussed in \autoref{sec:background}, we study the effect of learning rates on image-level, SUT-level, and human-level metrics. Experiments are conducted on all three tasks (\imagenet, \celeba, \yolo) using a sweep over nine \(\mathrm{lr}_{\min}\) and \(\mathrm{lr}_{\max}\) configurations (described later). As RQ\textsubscript{4} focuses on overall convergence behavior and runtime trends rather than class-wise statistical comparisons, we collect two test cases per class and configuration, resulting in 20 test cases per learning rate schedule and yielding \(9 \times 10 \times 2 = 180\) test cases per task and 540 in total. We intentionally limit the number of samples per class, as extremely small learning rates substantially increase optimization runtime while not affecting the overall trends analyzed in this research question. For each approach, we record images, runtime, and either human quality ratings (on \imagenet only) or FID scores as a surrogate for realism on other datasets. Annotators evaluate realism on a 1--5 scale and are assigned a single class, receiving 20 images and one attention-check question per batch. Our methodology adapts the weights of a HyperNet placed on top of a selected diffusion model; because we backpropagate through the diffusion model to generate failure-inducing perturbations for the SUT, the associated hyperparameters must be carefully balanced. We therefore use a OneCycle learning-rate schedule together with the AdamW optimizer (\autoref{sec:optimizer}), which reduces manual tuning. The schedule still requires selecting a minimal learning rate $lr_{\min}$ and a maximal rate $lr_{\max} = 100 \cdot lr_{\min}$, and we therefore sweep a predefined range of minimal learning rates. For the \imagenet task, this sweep consists of nine configurations defined by progressively increasing $lr_{\min}$, with $lr_{\max}$ fixed by the $100\times$ ratio.

To assess robustness to stochasticity, we repeat the \tool experiments using five additional random seeds ({1, 9127, 23481, 48673, 938124}) in addition to the default seed. This yields six independent distributions of 100 test cases each, which we compare using kernel density estimation and the Anderson--Darling k-sample test.

\[
\mathcal{LR}_{\min} = \{\, m \cdot 10^{-k} \mid m \in \{1,3\},\; k \in  \{8,7,6,5,4\} \,\}.
\]

For the \celeba task, we define the learning-rate range to alternate between $3e^{-a} \rightarrow 1e^{-b}$ and $1e^{-a} \rightarrow 4e^{-b}$, where $a$ decreases from 8 to 4 and $b$ decreases from 6 to 4. For the \yolo task, the range alternates between $5e^{-a} \rightarrow 8e^{-b}$ and $1e^{-a} \rightarrow 4e^{-b}$, with $a$ decreasing from 6 to 2 and $b$ decreasing from 5 to 1.

\subsection{Results}

\subsubsection{Effectiveness (RQ\textsubscript{1})}
\input{tables/rq1_merged}

To assess the effectiveness of \tool, we evaluate both image-based and SUT-based performance metrics (\autoref{sec:metrics}).
For the classification task, both \tool and \mrm achieve a misclassification rate of $1$ for all SUTs, meaning all generated test cases successfully induce failures in the SUT. \mimicry performs worse with a rate of $0.82-0.95$. Regarding the escape ratio, \tool achieves a perfect value of $0$, indicating that all generated test cases trigger their targeted misbehavior, which is an effect of the loss terms, steering towards specific targets.
In contrast, both baselines exhibit substantially higher escape ratios, with \mrm exceeding $0.75$ and \mimicry ranging from $0.12-0.52$.

For the binary classification task, \tool again achieves a misclassification rate of $1$, dominating the baselines, demonstrating consistent effectiveness. \mimicry struggles due to the absence of clear target classes in this task. While \mimicry's escape ratio is lower than \tool's, this reduction stems from the fact that \mimicry struggles to do any meaningful reduction of confidences, rather than having more fine-grained control.

For the object detection task, \tool again achieves a misclassification rate of \(1\). The accompanying confidence reductions across the detected objects indicate that the method consistently suppresses model confidence, even when evaluation is restricted to the top-5 detections. This demonstrates reliable control at the detection level. In contrast, \mrm shows weaker effectiveness: latent-space noise perturbations reduce only \(\sim 0.72\) of detections on average.
The corresponding mean confidence reduction for the relevant classes is also lower, with \mrm achieving \(0.83\) compared to \(0.95\) for \tool.

Looking at the image-based metrics, across all tasks, \tool surpasses the baselines in MS-SSIM, the embedding diversity of generated test cases, and the trace difference between original–target pairs. The only exception is \yolo, where \mrm has a higher target embedding diversity.

\begin{tcolorbox}[boxrule=0pt,sharp corners,boxsep=2pt,left=2pt,right=2pt,top=2.5pt,bottom=2pt]
\begin{center}
\begin{minipage}[t]{0.99\linewidth}
\textbf{RQ\textsubscript{1} (effectiveness):}  
\textit{\tool consistently induces failures across all tasks and SUTs (misclassification rate $=1$), maintains better control over targeted misbehavior (low escape ratio), and produces test cases that remain closer to the data distribution while achieving higher structural similarity and more stable embedding behavior than the baselines.}
\end{minipage}
\end{center}
\end{tcolorbox}

\subsubsection{Efficiency (RQ\textsubscript{2})}

\noindent\input{tables/rq2_runt}\noindent As shown in \autoref{tab:runt}, \tool consistently uses the lowest budget across tasks.
On \imagenet, \tool uses an average budget of $\sim 25$, which is only about $1\%$ of \mimicry's budget ($\sim2500$). Compared to \mrm, \tool requires roughly $28\times$ fewer evaluations. The standard deviations of \tool and \mrm are relatively high compared to their means, while \mimicry shows much lower relative variance. A similar trend holds for \celeba, where \mrm is not included.

For the \yolo task, \tool's budget usage is even lower at $\sim6$, compared to the $\sim25{-}30$ evaluations seen for \imagenet and \celeba, and considerably lower than the $\sim232$ of \mrm in the same task. 

Budget usage alone does not show efficiency as runtime also depends on the generative model architecture. On \imagenet, \tool achieves the lowest mean runtime overall, also reflected by the TFLOPs per test case. \mimicry shows a comparable mean runtime but with substantially lower variance, however substantially higher TFLOPs measure. \mrm is the slowest baseline, requiring roughly $2\times$ the runtime of \tool on average. For \celeba, \tool is slower than \mimicry, with mean runtime of $\sim220$ seconds and $\sim52$ seconds, respectively. In this setting, the lower runtime of \mimicry is primarily attributable to its substantially smaller generative models, as reflected in the TFLOPs measure, rather than differences in optimization budget usage.
Looking at the TFLOPs measure for all methods, we observe that, for every task except \celeba, \tool has substantially lower computational requirements than the baselines, while the runtime does not decrease proportionally. This discrepancy is due to the fact that \tool backpropagates through the diffusion process, which involves relatively few FLOPs but remains computationally expensive.

Finally, extrapolating \mimicry's budget usage using \tool's diffusion-based runtime suggests that a diffusion-augmented \mimicry would incur runtime roughly $5{-}10\times$ higher than \tool. This indicates that directly leveraging gradients, as \tool does, provides a substantial efficiency advantage.

\begin{tcolorbox}[boxrule=0pt,sharp corners,boxsep=2pt,left=2pt,right=2pt,top=2.5pt,bottom=2pt]
\begin{center}
\begin{minipage}[t]{0.99\linewidth}
\textbf{RQ\textsubscript{2} (efficiency)}: \textit{Overall, \tool is substantially more evaluation-efficient than both \mimicry and \mrm, requiring orders of magnitude fewer SUT evaluations. Competing approaches incur up to two orders of magnitude higher evaluation budgets and TFLOPs, while runtime efficiency varies by task and model architecture. In particular, although \tool is not always the fastest in wall-clock time, extrapolating diffusion-based runtimes to the evaluation budgets of \mimicry would result in a $5{-}10\times$ higher overall runtime, confirming \tool's better efficiency–runtime trade-off.}
\end{minipage}
\end{center}
\end{tcolorbox}

\subsubsection{Validity (RQ\textsubscript{3})}

Across the two batches used in the human assessment, we obtained 11 and 9 valid responses, respectively. The inter-rater agreement, measured using Fleiss' kappa, was $0.336 \pm 0.053$. While there is no universally agreed-upon interpretation scale, values between $0.2$ and $0.4$ are commonly considered to indicate a fair level of agreement~\cite{Landis:1977}.

\input{tables/rq3_im}

Looking at \autoref{tab:rq3_im}, we observe that \tool outperforms the two baselines. In terms of label preservation, \tool more than doubles the performance of \mimicry and achieves roughly $60\%$ higher preservation than \mrm. A similar trend is visible for Image Realism, where \tool performs approximately $80\%$ better than \mimicry and again around $60\%$ better than \mrm. Furthermore, \tool produces fewer ambiguous cases, indicating that its outputs are easier for human raters to judge and suggesting higher semantic validity overall.

\input{tables/rq3_im_stat}

To investigate whether these improvements are statistically significant, we compute $p$-values using the Wilcoxon signed-rank test~\cite{Wilcoxon1945}
for which $p<0.05$ is the threshold for significance, and effect sizes using Cohen's $d$, where values above $0.5$ indicate medium effects and values above $0.8$ indicate large effects. As shown in \autoref{tab:rq3_im_stat}, \tool significantly outperforms both baselines in both label preservation and image realism, with medium to large effect sizes.
\input{figures/small_comp}
These quantitative findings are supported by the qualitative examples.

In \autoref{fig:im_small_comp}, results produced by \tool remain visually close to the original images, confirming that it maintains label semantics and generates realistic, coherent perturbations. In contrast, \mrm introduces coarser and more disruptive changes, reducing semantic validity and increasing the likelihood that annotators perceive objects as missing or corrupted.

This tendency is even stronger in examples from \mimicry, which often lose structural and semantic integrity. While some generated images resemble the general class, many examples diverge significantly from the origin in terms of spatial layout and meaningful features. This behavior is a known limitation of GAN-based architectures, which can struggle to preserve fine-grained structure. The larger discrepancies between origin and result images further illustrate that \mimicry provides limited control over the manipulation process

\begin{figure}
    \centering
    \includegraphics[width=0.5\columnwidth]{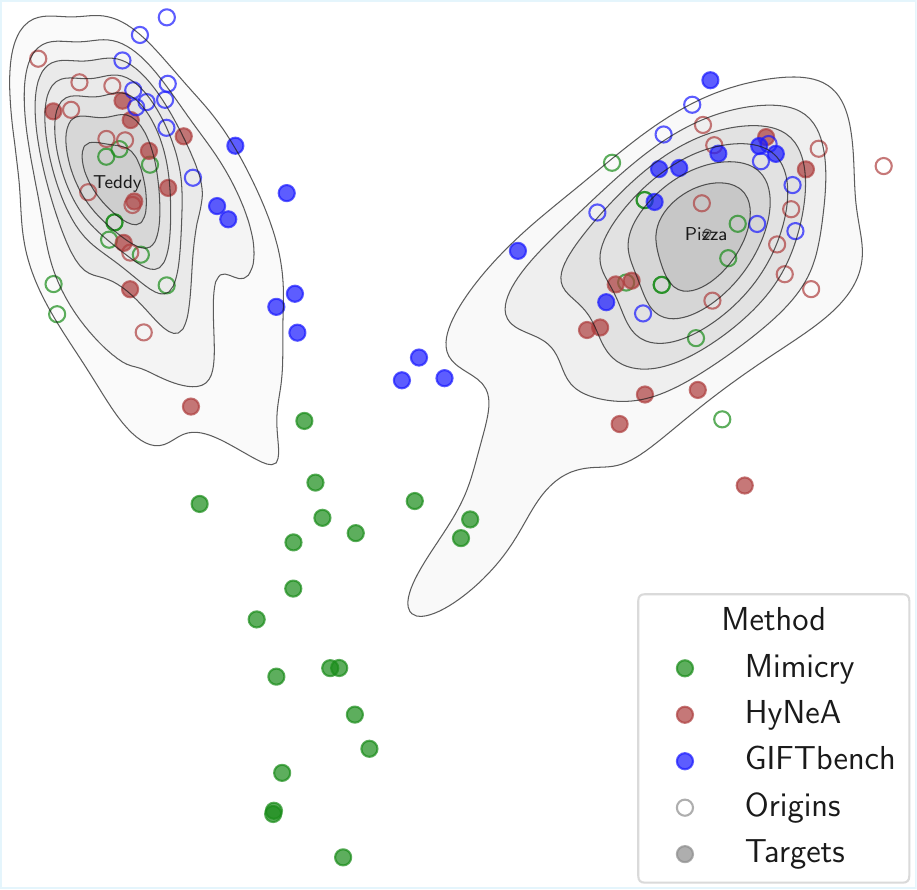}
    \caption{Embedding comparison for ''pizza'' and ''teddy bear'' across methods.}
    \label{fig:emb}
\end{figure}

\autoref{fig:emb} provides a qualitative complement to the human assessment. It shows how generated samples relate to the ImageNet validation-set embedding distribution for the ''pizza'' and ''teddy bear'' classes.
Although embedding distances do not directly measure semantic validity, larger deviations tend to coincide with images that annotators judged as less realistic or semantically unclear.
Samples generated by \mimicry and \mrm generally show larger shifts in embedding space, consistent with the lower label preservation and realism scores observed in the human evaluation. In contrast, \tool typically exhibits smaller embedding shifts, aligning with annotators' assessments that these images better preserve visual structure and class semantics. This observation is correlational and intended to support, rather than replace, the qualitative and human-evaluation results.
\input{figures/small_comp_ca}
For the \celeba task (\autoref{fig:ca_small_comp}), \tool again demonstrates a more fine-grained and controlled behavior. Unlike \mimicry, which requires interpolation towards a fixed latent target, \tool directly manipulates the image while preserving the core identity and structure. This property is essential for effective test-case generation, as seen in the binary attribute states shown in the figures. The third column, which visualizes the total change between origin and result, highlights that \tool modifies only local or attribute-relevant regions, while \mimicry often introduces global changes that overwhelm the original image. This makes \mimicry unsuitable for tasks where no reliable conditional target exists.

\input{figures/small_comp_yolo}
Finally, the \yolo task in \autoref{fig:yolo_small_comp} illustrates how \tool suppresses detections of selected classes while keeping the overall scene semantically intact. Bounding boxes for the top-5 detections are shown only when their confidence exceeds \(0.5\), preventing low-certainty predictions from appearing in the visualization. In these examples, \tool maintains scene structure and visual coherence, modifying only the image regions necessary to reduce detector confidence. In contrast, \mrm primarily degrades images until they fall sufficiently out of distribution for the SUT to function reliably.
More examples of images generated for the three tasks can be found in the Appendix.

\begin{tcolorbox}[boxrule=0pt,sharp corners,boxsep=2pt,left=2pt,right=2pt,top=2.5pt,bottom=2pt]
\begin{center}
\begin{minipage}[t]{0.99\linewidth}
\textbf{RQ\textsubscript{3} (validity)}:
\textit{Human evaluation on \imagenet shows that \tool produces higher-quality and more semantically meaningful manipulations than both \mimicry and \mrm. \tool more than doubles \mimicry's label-preservation rate, achieves about $60\%$ higher preservation than \mrm, and delivers roughly the same improvement in perceived realism. Qualitative assessments align with these results: \tool better preserves object identity and structure, and restricts changes to semantically relevant regions, while the baselines introduce broader, less controlled modifications.}
\end{minipage}
\end{center}
\end{tcolorbox}

\subsubsection{Sensitivity (RQ\textsubscript{4})}\label{sec:sens}
\begin{figure}[t]
\centering
\begin{minipage}{0.52\textwidth}
    \centering
    \includegraphics[width=\linewidth]{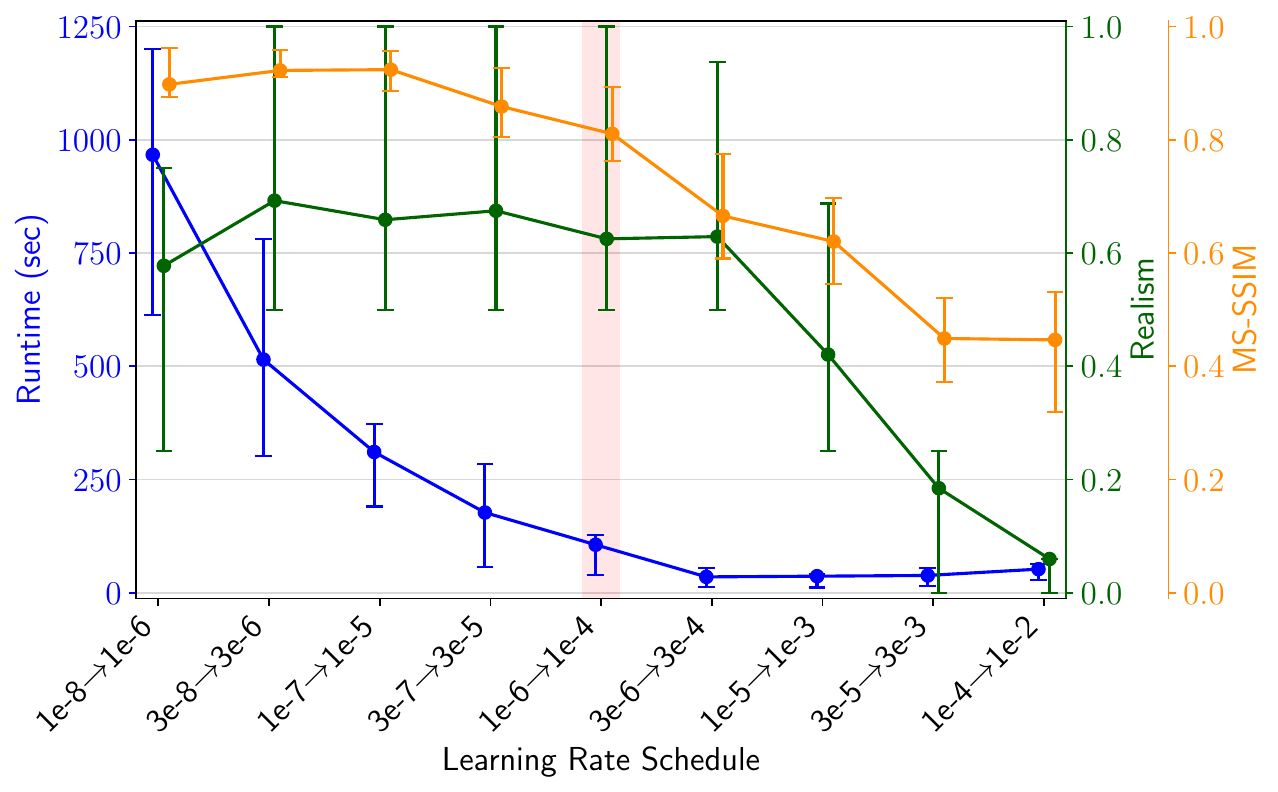}
    \caption{Runtime–quality trade-off across learning-rate schedules (\imagenet[icon]).}
    \label{fig:im_sweep}
\end{minipage}
\hfill
\begin{minipage}{0.47\textwidth}
    \centering
    \includegraphics[width=\linewidth]{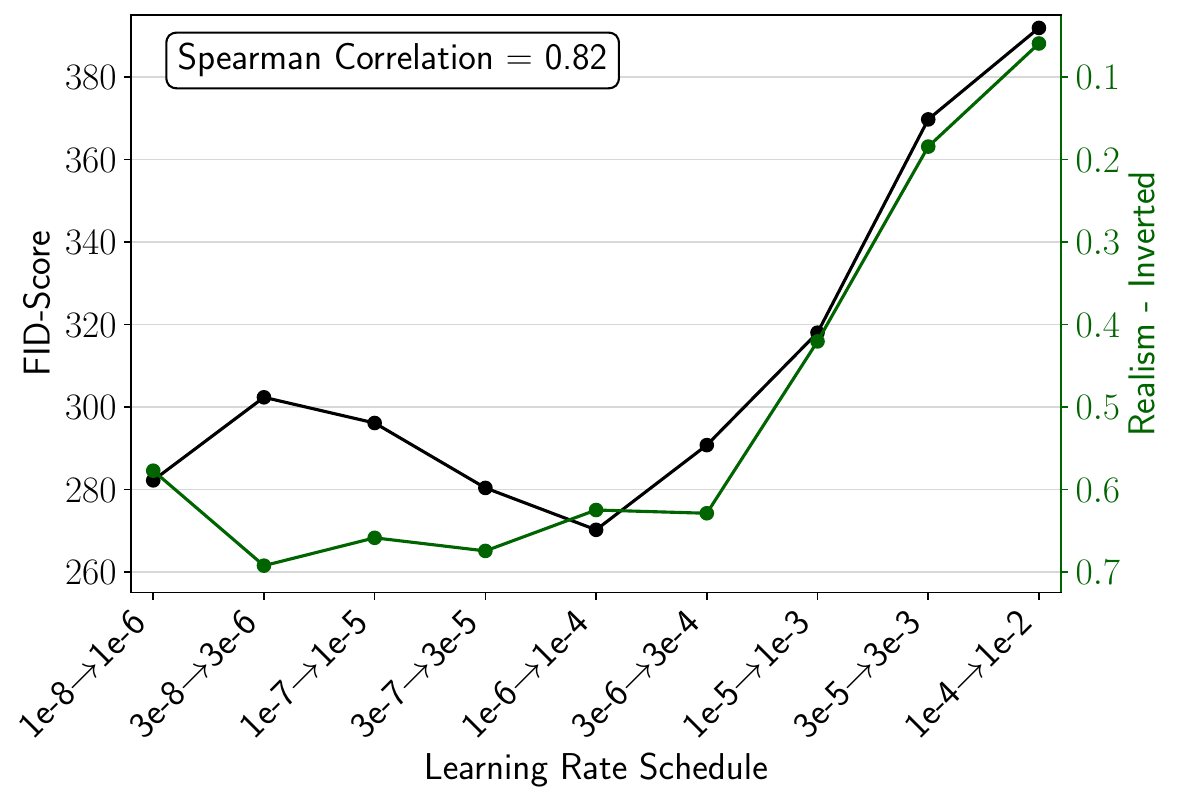}
    \caption{Correlation between FID and Human annotated Image-Realism.}
    \label{fig:sweep_corr}
\end{minipage}
\end{figure}

\begin{figure}[b]
    \centering
    \includegraphics[width=\linewidth]{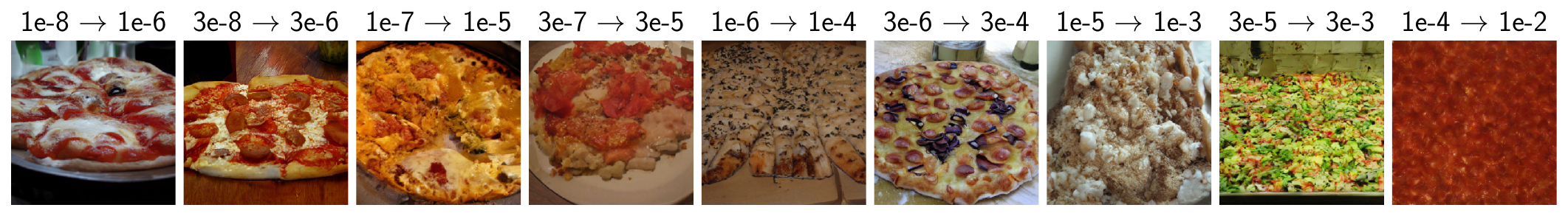}
    \caption{Examples of \tool outputs for fooling the SUT with initial class "pizza" across lr-schedules.}
    \label{fig:exlr}
\end{figure}

\autoref{fig:im_sweep} shows runtime in seconds \textcolor{blue}{(blue)} and perceived output quality \textcolor{darkgreen}{(green)}, where 1 denotes the highest quality and 0 the lowest. The shaded region marks the learning-rate range used in RQ\textsubscript{1}–RQ\textsubscript{3}. The error bars indicate the 25\textsuperscript{th} and 75\textsuperscript{th} quartile. 
Perceived output quality is measured through a human evaluation study, yielding \(6.3 \pm 1.8\) valid responses on average, with a validity rate of \(71.6\%\) across classes.

Very small learning rates lead to substantially longer runtime, in some cases up to 20 minutes per test case. Increasing the learning rate reduces runtime rapidly, stabilizing below roughly 200 seconds. Output quality shows the opposite trend: larger learning rates yield more aggressive perturbations and noticeably degrade visual fidelity. Learning rates below $10^{-6}$ do not substantially improve quality, indicating diminishing returns at the lower end. A value of $10^{-6}$ offers a practical balance for this task (see examples in  \autoref{fig:exlr}).

To assess structural change in the manipulated images, \autoref{fig:im_sweep} reports MS-SSIM values in \textcolor{orange}{orange}. Learning rates below $10^{-6}$ produce similar MS-SSIM values, whereas higher learning rates reduce MS-SSIM more quickly, indicating more pronounced structural changes.

As we aim to automate evaluation to the extent possible, we investigate whether human-annotated realism correlates with the widely-used FID score, which compares distributions of images. We compare the distribution of generated images produced by \tool to the distribution in the test or validation splits of the respective datasets. Because a lower FID score is generally better, we invert the human realism measure for comparison in \autoref{fig:sweep_corr}. As the figure shows, the general trend is highly correlated, which is confirmed by the Spearman correlation coefficient of $0.82$. This indicates that FID can serve as a reliable surrogate metric for the trade-off analyses used in the remaining datasets.

\begin{figure}[t]
  \centering
  \begin{subfigure}[b]{0.48\textwidth}
    \centering
    \includegraphics[width=\textwidth]{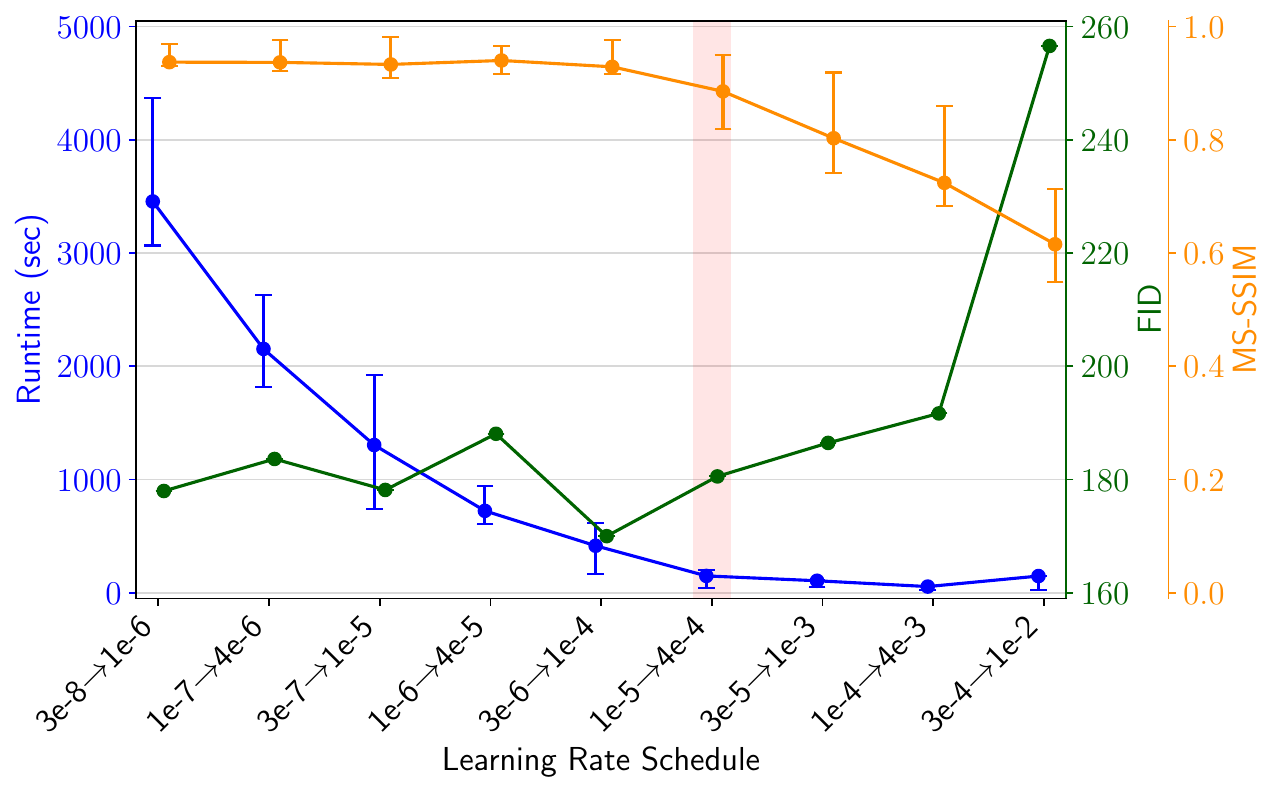}
    \caption{Tradeoff for \celeba[icon].}
    \label{fig:c_sweep}
  \end{subfigure}
  \hfill
  \begin{subfigure}[b]{0.48\textwidth}
    \centering
    \includegraphics[width=\textwidth]{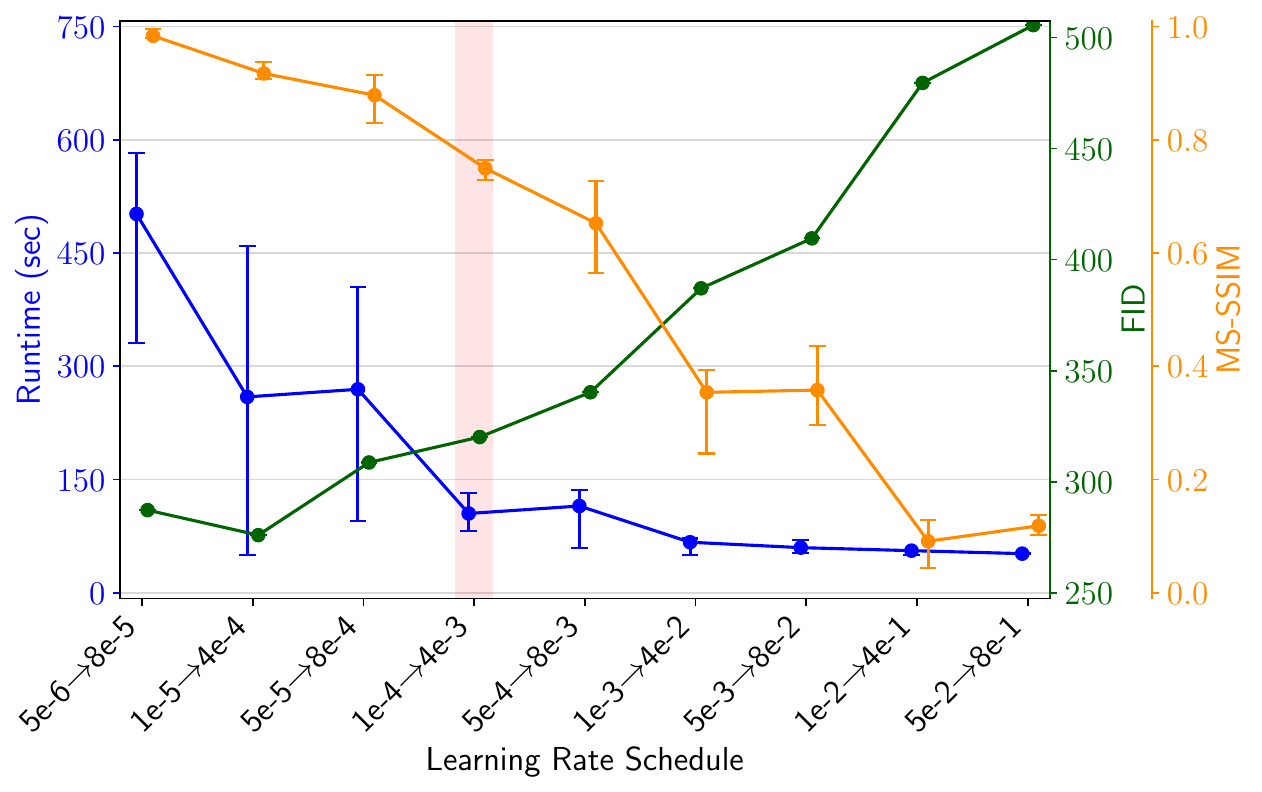}
    \caption{Tradeoff for \yolo[icon] Dataset.}
    \label{fig:y_sweep}
  \end{subfigure}
  \caption{Runtime–quality trade-off across learning-rate schedules.}
  \label{fig:dual_sweep}
\end{figure}

As shown in \autoref{fig:dual_sweep}, the red highlighted regions correspond to the schedules used in the final experiments for the remaining RQs. Generally, we aim for minimal degradation of realism (FID) and structural content (MS-SSIM), while keeping runtime low to facilitate efficient experimentation. For \celeba, the schedule ranging from $1\mathrm{e}{-5}\rightarrow 4\mathrm{e}{-4}$ provides the most appropriate trade-off across these aspects. For the Driving dataset, the learning rates are higher due to the type of data and the characteristics of the SUT, for which the most suitable schedule was found to be $1\mathrm{e}{-4}\rightarrow 4\mathrm{e}{-3}$.

\autoref{fig:seed-sensitivity} shows that the distributions of SSIM, LPIPS, and Runtime remain highly consistent across the evaluated random seeds. The kernel density estimates largely overlap, and the means obtained for the additional seed configurations (blue markers) all fall within the 99\% confidence interval of the default experiment (gray region), whose mean is indicated by the black dashed line. Anderson--Darling k-sample tests likewise found no evidence of statistically significant differences between seed-specific distributions ((p=0.85), (p=0.80), and (p=0.81) for SSIM, LPIPS, and Runtime, respectively). Together, these results indicate that \tool exhibits low sensitivity to random seed selection.

\begin{tcolorbox}[boxrule=0pt,sharp corners,boxsep=2pt,left=2pt,right=2pt,top=2.5pt,bottom=2pt]
\begin{center}
\begin{minipage}[t]{0.99\linewidth}
\textbf{RQ\textsubscript{4} (sensitivity):} \textit{Small learning rates substantially increase runtime without providing meaningful improvements in output quality or structural similarity. Larger learning rates significantly reduce runtime but introduce stronger, less controlled perturbations. The strong correlation between human-rated realism and FID enables automated evaluation. Across \imagenet, \celeba, and the Driving dataset, we show that it is possible to identify learning-rate schedules that strike the best balance between realism, structural preservation, and runtime efficiency, ensuring that \tool remains both effective and practical for large-scale experimentation. Furthermore, the observed results are robust to random seed selection, with no statistically significant differences in metric distributions across evaluated seeds.}
\end{minipage}
\end{center}
\end{tcolorbox}

\subsection{Threats to Validity}
\input{figures/sensitivity}
\subsubsection{Internal validity}
All methods were executed under the same computational budget, defined by the total number of SUT evaluations. Early termination conditions were applied consistently across methods to ensure comparable efficiency measurements. The generative models used in our experiments were taken from prior work and selected for their demonstrated generation quality~\cite{leng2025repa, rombach2022high}. All SUTs are publicly available through \texttt{torchvision}~\cite{torchvision2016}, \texttt{ultralytics}~\cite{Jocher_Ultralytics_YOLO_2023}, or replication packages~\cite{2026-Chen-DETECT}, supporting reproducibility.

For human evaluations, the survey structure was validated via a pilot run and attention-check questions ensured response quality~\cite{sorokin2008utility}. While identical starting conditions cannot be enforced due to sampling, multiple repetitions per test case mitigate variability. Human annotation is further constrained by budget ($\sim250\$$), creating a trade-off between the number of responses and fair compensation; we prioritized fewer high-quality responses to maximize annotation reliability.

\subsubsection{External validity}
Our evaluation considers three different testing tasks and a diverse set of SUTs, but these do not represent the full space of possible systems. Accordingly, generalization to other architectures or domains remains an open question. In particular, our empirical claims are limited to vision-based DL systems; extending HyNeA to language, audio, code, or multimodal systems would require modality-specific generators, control mechanisms, objectives, validity oracles, baselines, and metrics. The choice of generative model also affects external validity: more capable models tend to enable more effective test generation. Although we selected the strongest open-source models available for each task, results may differ when applied to weaker generative backbones.

%% file: tables/rq1_merged.tex
\begin{table}[t]
\caption{RQ\textsubscript{1}: Effectiveness results for each approach and tasks.}
\resizebox{\textwidth}{!}{
\begin{tabular}{lllccccccc}\toprule
& & & \multicolumn{3}{c}{Task Performance} & \multicolumn{4}{c}{Image Metrics} \\
\cmidrule(lr){4-6}\cmidrule(lr){7-10}
Task & Model & Method
& Misclass Rate $\uparrow$ 
& Escape Ratio $\downarrow$ 
& Conf. Red. $\uparrow$
& MS-SSIM $\uparrow$ 
& LPIPS $\downarrow$ 
& Diversity T $\uparrow$ 
& Trace Diff $\downarrow$ \\
\midrule

\multirow{9}{*}{\textit{\imagenet[icon]}}

& \multirow{3}{*}{\textit{WRN}}
& \tool
& \textbf{1.00} & \textbf{0.00} & \multirow{3}{*}{-} 
& \textbf{0.760 $\pm$ 0.126} & \textbf{0.318 $\pm$ 0.233} & \textbf{0.179} & \textbf{79.60} \\

& & \mimicry
& 0.82 & 0.52 &  
& 0.213 $\pm$ 0.115 & 0.732 $\pm$ 0.062 & 0.147 & 197.21 \\

& & \mrm
& \textbf{1.00} & 0.84 &  
& 0.563 $\pm$ 0.086 & 0.499 $\pm$ 0.068 & 0.102 & 119.92 \\

\cmidrule(lr){2-10}

& \multirow{3}{*}{\textit{EN2}}
& \tool
& \textbf{1.00} & \textbf{0.00} & \multirow{3}{*}{-} 
& \textbf{0.739 $\pm$ 0.151} & \textbf{0.346 $\pm$ 0.148} & \textbf{0.190} & \textbf{81.23}  \\

& & \mimicry
& 0.95 & 0.12 &  
& 0.213 $\pm$ 0.111 & 0.730  $\pm$ 0.062 & 0.149 & 199.95 \\

& & \mrm
& \textbf{1.00} & 0.84 &  
& 0.565 $\pm$ 0.105 & 0.548 $\pm$ 0.107 & 0.099 & 136.74 \\

\cmidrule(lr){2-10}

& \multirow{3}{*}{\textit{ViT}}
& \tool
& \textbf{1.00} & \textbf{0.00} & \multirow{3}{*}{-} 
& \textbf{0.781 $\pm$ 0.122} & \textbf{0.283 $\pm$ 0.108} & \textbf{0.184} & \textbf{78.25} \\

& & \mimicry
& 0.90 & 0.25 &  
& 0.193 $\pm$ 0.110 & 0.730  $\pm$ 0.069 & 0.151 & 186.01    \\

& & \mrm
& \textbf{1.00} & 0.75 &  
& 0.350 $\pm$ 0.074 & 0.592 $\pm$ 0.082 & 0.130 & 126.94 \\

\midrule

\multirow{3}{*}{\textit{\celeba[icon]}}
& \multirow{3}{*}{RN50}

& \tool
& \textbf{1.00} & 0.11 & \multirow{3}{*}{-} 
& \textbf{0.904 $\pm$ 0.069} & \textbf{0.241 $\pm$ 0.143} & \textbf{0.068} & \textbf{32.80} \\

& & \mimicry
& 0.01 & \textbf{0.01} &  
& 0.491 $\pm$ 0.120 & 0.590 $\pm$ 0.068 & 0.052 & 70.77 \\

& & \mrm
& 0.47 & 0.08 &  
& 0.758 $\pm$ 0.080 & 0.389 $\pm$ 0.056 & 0.047 & 36.23 \\

\midrule

\multirow{2}{*}{\textit{\yolo[icon]}}
& \multirow{2}{*}{YoloV8}

& \tool
& \textbf{1.00 $\pm$ 0.00} & \multirow{2}{*}{-} & \textbf{0.95 $\pm$ 0.11}
& \textbf{0.736 $\pm$ 0.177} & \textbf{0.310 $\pm$ 0.156} & 0.094 & \textbf{60.71} \\

& & \mrm
& 0.72 $\pm$ 0.45 &  & 0.83 $\pm$ 0.25
& 0.606 $\pm$ 0.266 & 0.454 $\pm$ 0.153 & \textbf{0.102} & 191.50 \\

\bottomrule
\end{tabular}\label{tab:merged_rq1}}
\end{table}

%% file: tables/rq2_runt.tex
\begin{table}[t]
\footnotesize
\caption{RQ\textsubscript{2}: efficiency results of each approach.}
\begin{tabular}{llllll}\toprule
& & Method & Runtime (sec) $\downarrow$ & Budget Used $\downarrow$ & TFLOPs $\downarrow$\\
\midrule

\multirow{10}{*}{\textit{\imagenet[icon]}}

& \multirow{3}{*}{\textit{WRN}}
& \tool
& \textbf{94.41 $\pm$ 99.26} & \textbf{25.29 $\pm$ 26.72} & \textbf{858 $\pm$ 901} \\

& & \mimicry
& 108.53 $\pm$ 16.93 & 2498.16 $\pm$ 390.57 & 3739 $\pm$ 582 \\

& & \mrm
& 205.74 $\pm$ 198.99 & 699.63 $\pm$ 748.59 & 7387 $\pm$ 7693 \\

\cmidrule(lr){2-6}

& \multirow{3}{*}{\textit{EN2}}
& \tool
& \textbf{98.38 $\pm$ 109.45} & \textbf{25.48 $\pm$ 28.30} & \textbf{864 $\pm$ 954} \\

& & \mimicry
& 110.01 $\pm$ 16.12 & 2485.19 $\pm$ 370.40 & 3719.6 $\pm$ 551.6 \\

& & \mrm
& 190.39 $\pm$ 149.50 & 282.00 $\pm$ 221.05 & 2978 $\pm$ 2214 \\

\cmidrule(lr){2-6}

& \multirow{3}{*}{\textit{ViT}}
& \tool
& \textbf{64.07 $\pm$ 66.16} & \textbf{16.55 $\pm$ 17.05} & \textbf{561 $\pm$ 575} \\

& & \mimicry
& 120.80 $\pm$ 14.02 & 2529.10 $\pm$ 291.56 & 3785.3 $\pm$ 434.2 \\

& & \mrm
& 212.70 $\pm$ 179.63 & 318.25 $\pm$ 267.82 & 3360 $\pm$ 2756 \\

\cmidrule(lr){2-6}

& average
& \mimicry~(Diffusion)
& $\sim1149$ & - & - \\

\midrule

\multirow{4}{*}{\textit{\celeba[icon]}}

& \multirow{4}{*}{Custom}

& \tool
& 220.89 $\pm$ 228.82 & \textbf{30.30 $\pm$ 31.43} & 852 $\pm$ 879 \\

& & \mimicry
& \textbf{51.87 $\pm$ 3.66} & 2673.20 $\pm$ 37.63 & \textbf{199 $\pm$ 3} \\

& & \mimicry~(Diffusion)
& $\sim1261$ & - & - \\

& & \mrm
& 1217.24 $\pm$ 735.71 & 1796.00 $\pm$ 1088.11 & 18963 $\pm$ 11296 \\

\midrule

\multirow{2}{*}{\textit{\yolo[icon]}}

& \multirow{2}{*}{YOLO}

& \tool
& \textbf{113.03 $\pm$ 111.70} & \textbf{5.71 $\pm$ 5.69} & \textbf{728.5 $\pm$ 718.0} \\

& & \mrm
& 174.58 $\pm$ 152.02 & 232.12 $\pm$ 245.72 & 13486.4 $\pm$ 14130.0 \\

\bottomrule
\end{tabular}\label{tab:runt}
\end{table}

%% file: tables/rq3_im.tex
\begin{table}[t]
\footnotesize
\caption{Comparison of approaches for quality measures (\textbf{bold} values indicate best value).}
\begin{tabular}{llll}\toprule
          & \textit{Label Preservation} $\uparrow$ & \textit{Image-Realism} $\uparrow$ & \textit{Ambiguity Cases} $\downarrow$ \\ \midrule
\tool    & \textbf{0.787 $\pm$ 0.401}     & \textbf{0.750 $\pm$ 0.237} & $\mathbf{2.5\%}$           \\
\mimicry   & 0.287 $\pm$ 0.431              & 0.388 $\pm$ 0.340          & $7.5\%$                  \\
\mrm & 0.463 $\pm$ 0.479              & 0.438 $\pm$ 0.361          & $7.5\%$                    \\\bottomrule
\end{tabular}\label{tab:rq3_im}
\end{table}

%% file: tables/rq3_im_stat.tex
\begin{table}[t]
\footnotesize
\caption{Statistical Analysis of Comparing \tool to baseline methods. (\textbf{bold} values are significant)}
\begin{tabular}{lllll} \toprule
             & \multicolumn{2}{l}{\textit{Label Preservation}} & \multicolumn{2}{l}{Image-Realism}    \\ 
\diagbox{\tool $>$}{Measure}& \textit{p-value}       & \textit{Cohen's d}      & \textit{p-value} & \textit{Cohen's d} \\ \midrule
\mimicry  & \textbf{9.6e-6}                & 1.185                 & \textbf{2.7e-3}          & 1.206            \\
\mrm &\textbf{8.8e-3}                & 0.726                 & \textbf{6.1e-3}          & 0.997            \\ \bottomrule
\end{tabular}\label{tab:rq3_im_stat}
\end{table}

%% file: figures/small_comp.tex
\begin{figure}[t]
    \centering
    \begin{subfigure}[t]{0.3\textwidth}
        \centering
        \includegraphics[width=\linewidth]{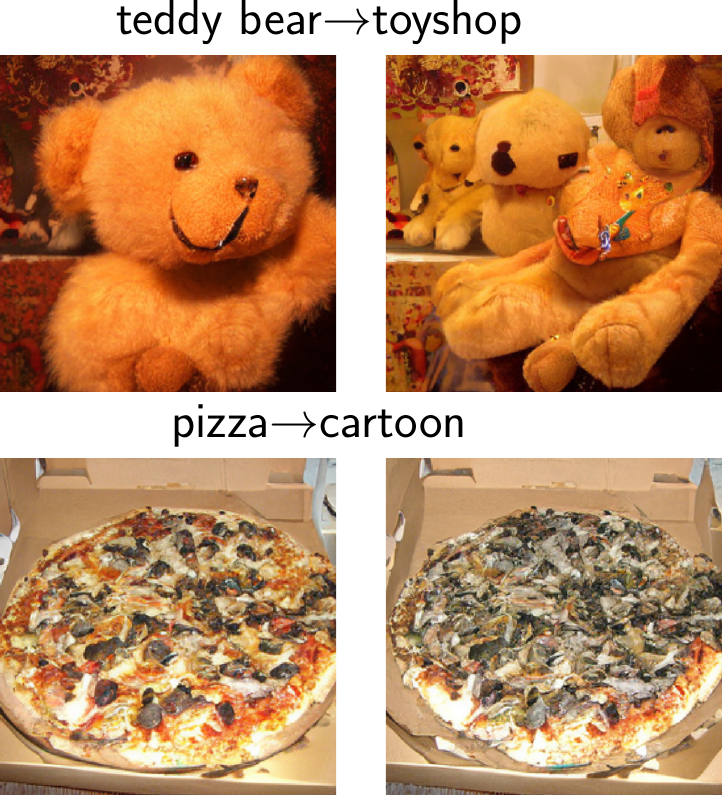}
        \caption{\tool}
    \end{subfigure}
    \hspace{0.02\textwidth}
    \begin{subfigure}[t]{0.3\textwidth}
        \centering
        \includegraphics[width=\linewidth]{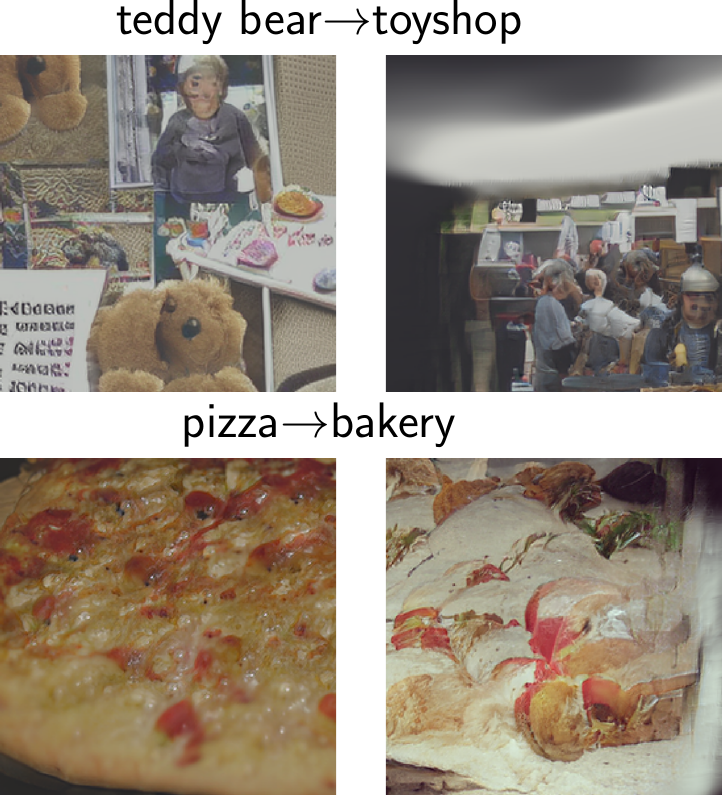}
        \caption{\mimicry}
    \end{subfigure}
    \hspace{0.02\textwidth}
    \begin{subfigure}[t]{0.3\textwidth}
        \centering
        \includegraphics[width=\linewidth]{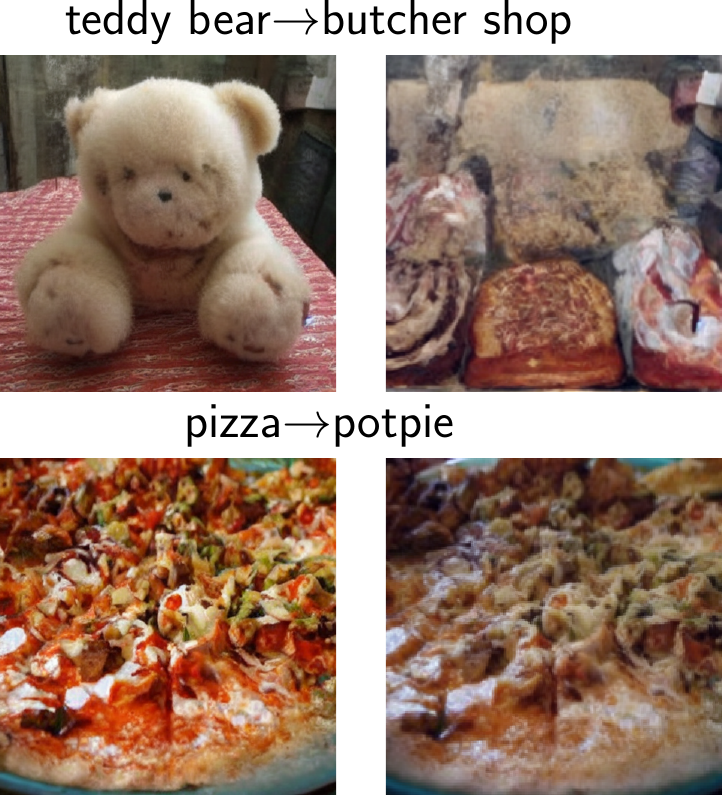}
        \caption{\mrm}
    \end{subfigure}

    \caption{Example Origins and Targets for \imagenet[icon]. Predicted Origin and Target class are shown on top.}
    \label{fig:im_small_comp}
\end{figure}

%% file: figures/small_comp_ca.tex
\begin{figure}[t]
    \centering
    \begin{subfigure}[t]{0.3\textwidth}
        \centering
        \includegraphics[width=\linewidth]{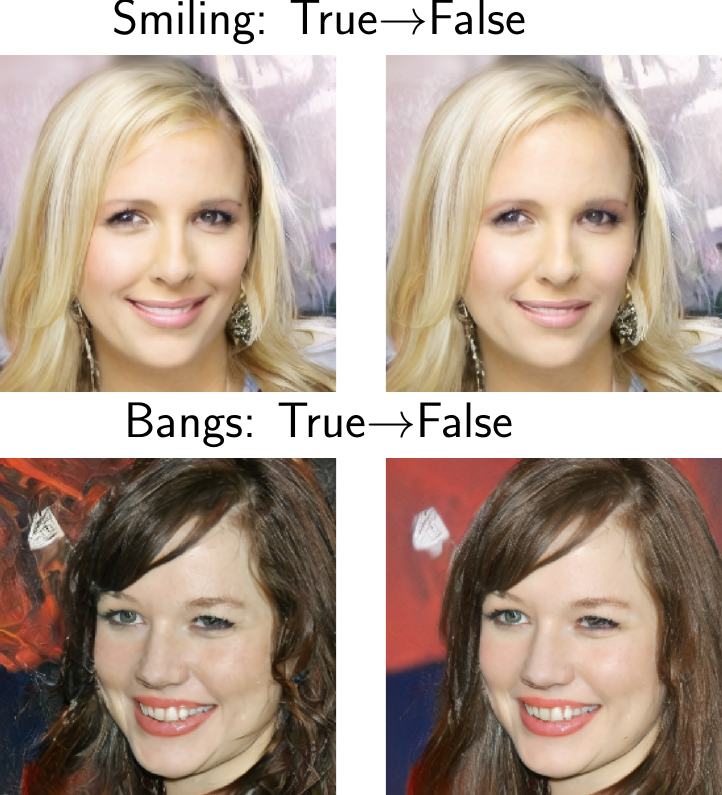}
        \caption{\tool}
    \end{subfigure}
    \hspace{0.02\textwidth}
    \begin{subfigure}[t]{0.3\textwidth}
        \centering
        \includegraphics[width=\linewidth]{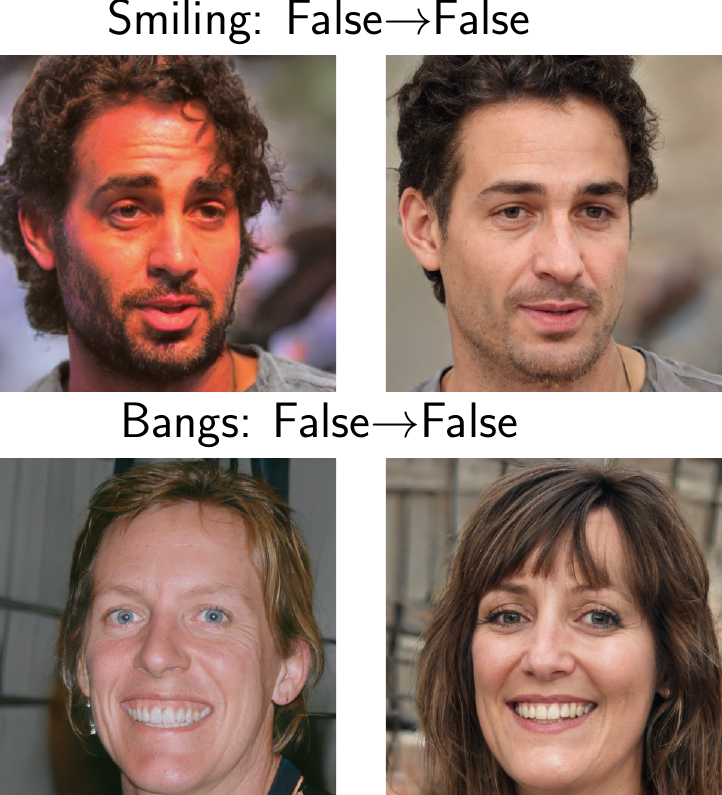}
        \caption{\mimicry}
    \end{subfigure}
    \hspace{0.02\textwidth}
    \begin{subfigure}[t]{0.3\textwidth}
        \centering
        \includegraphics[width=\linewidth]{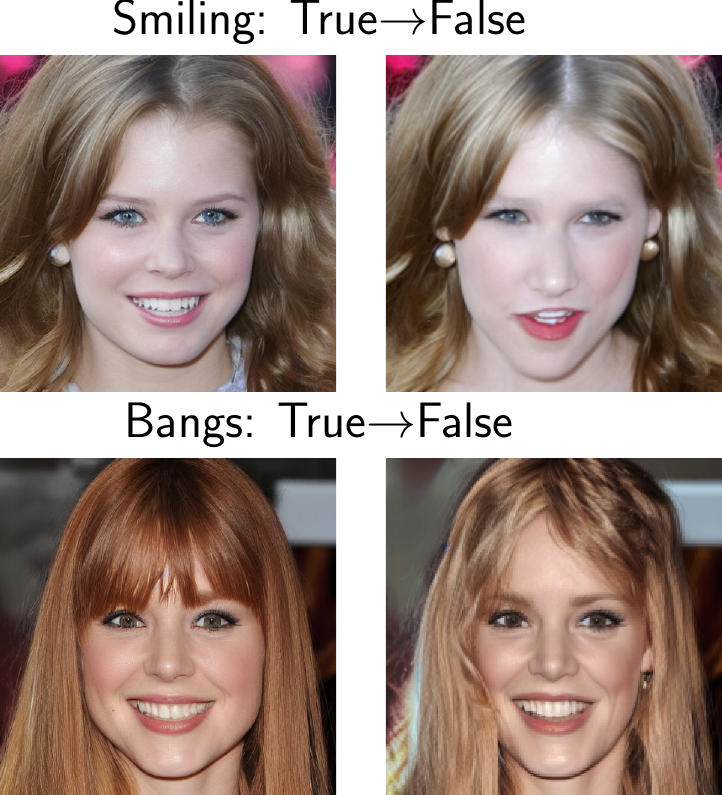}
        \caption{\mrm}
    \end{subfigure}

    \caption{Example Origins and Targets for \celeba[icon]. Change of predicted attribute presence on top.}
    \label{fig:ca_small_comp}
\end{figure}

%% file: figures/small_comp_yolo.tex
\begin{figure}[t]
    \centering
    \begin{subfigure}[t]{0.3\textwidth}
        \centering
        \includegraphics[width=\linewidth]{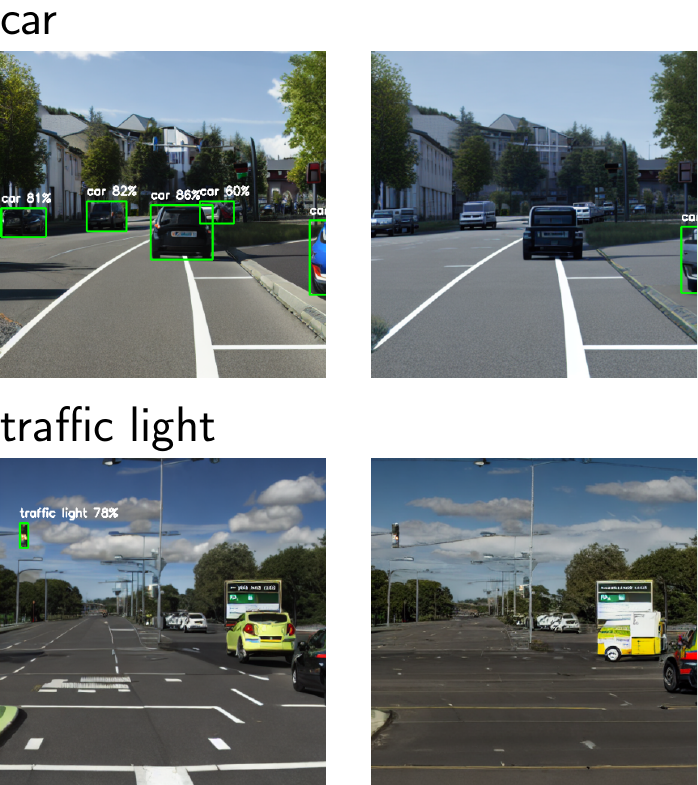}
        \caption{\tool}
    \end{subfigure}
    \hspace{0.02\textwidth}
    \begin{subfigure}[t]{0.3\textwidth}
        \centering
        \includegraphics[width=\linewidth]{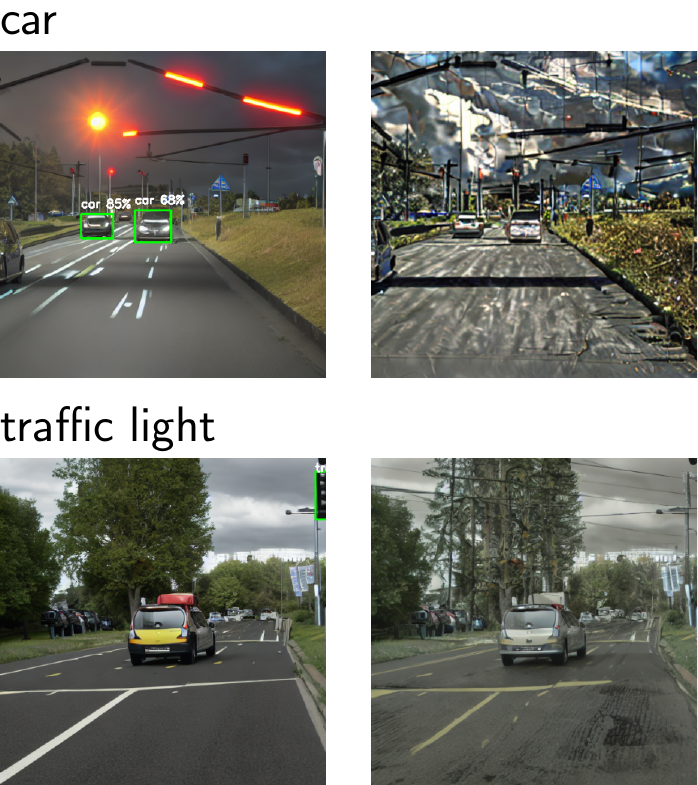}
        \caption{\mrm}
    \end{subfigure}

    \caption{Example Origins and Targets for \yolo[icon]. The targeted objects are shown above the images, with detected bounding boxes and their corresponding confidence scores displayed.}
    \label{fig:yolo_small_comp}
\end{figure}

%% file: figures/sensitivity.tex
\begin{figure*}[t]
    \centering

    \begin{subfigure}[t]{0.32\textwidth}
        \centering
        \includegraphics[width=\linewidth]{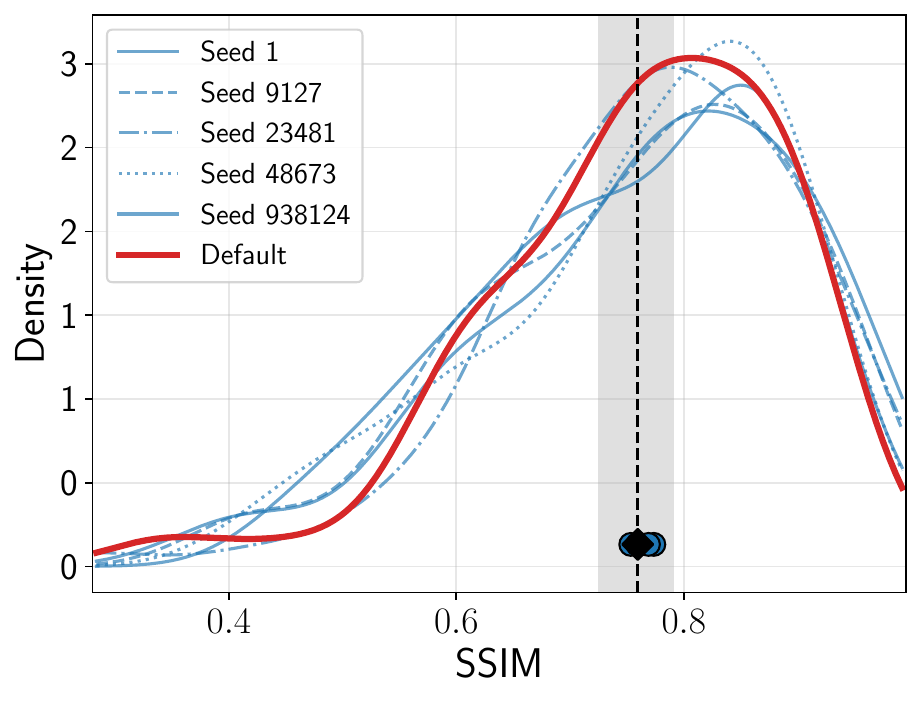}
        \caption{KDE of SSIM for different seeds.}
        \label{fig:seed-sens-ssim}
    \end{subfigure}
    \hfill
    \begin{subfigure}[t]{0.32\textwidth}
        \centering
        \includegraphics[width=\linewidth]{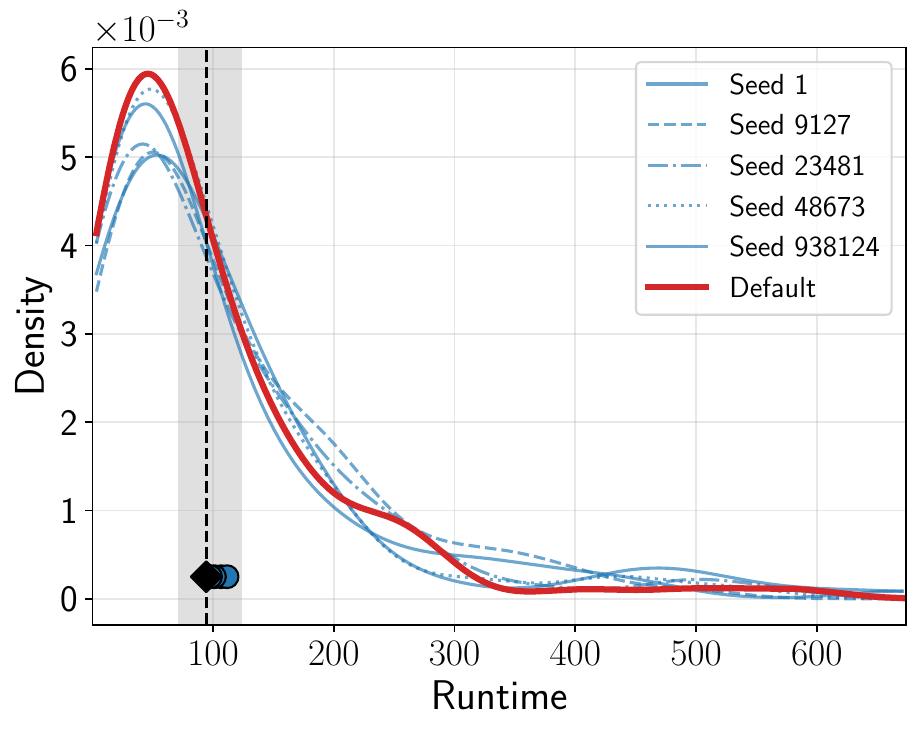}
        \caption{KDE of Runtime for different seeds.}
        \label{fig:seed-sens-runtime}
    \end{subfigure}
    \hfill
    \begin{subfigure}[t]{0.32\textwidth}
        \centering
        \includegraphics[width=\linewidth]{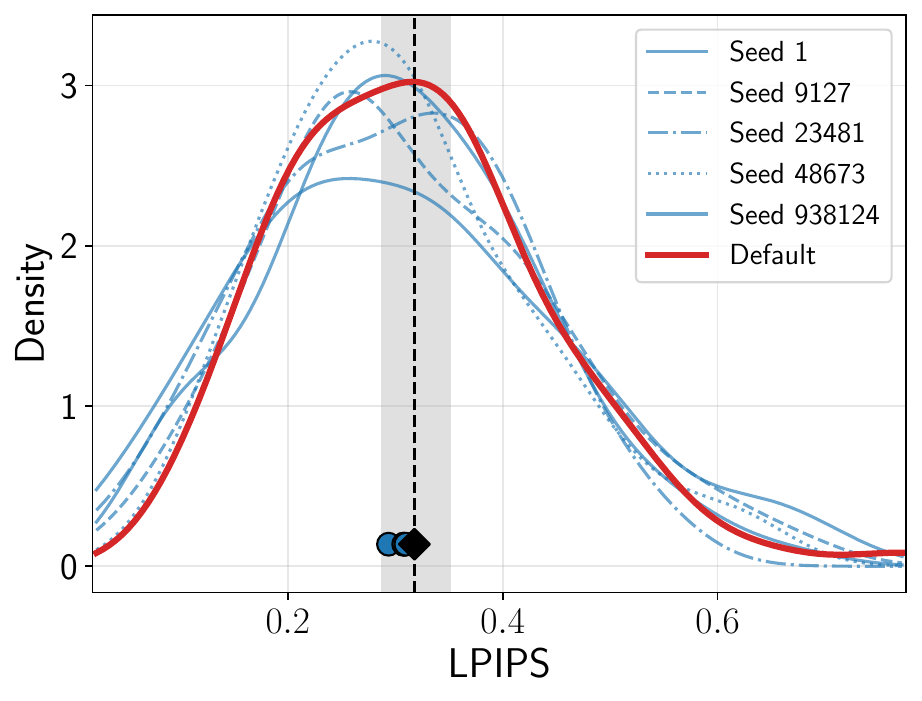}
        \caption{KDE of LPIPS for different seeds.}
        \label{fig:seed-sens-lpips}
    \end{subfigure}

    \caption{Blue lines show runs with alternative random seeds, while the red line represents the default experiment. The gray shaded region denotes the 99\% bootstrap confidence interval of the default mean, and the dashed black line marks the mean itself. Blue circles indicate the means of the alternative-seed runs, and the black diamond indicates the default mean.}
    \label{fig:seed-sensitivity}
\end{figure*}

%% file: sec/5_discussion.tex
\section{Discussion}

\head{Diffusion models substantially improve the quality of generated test cases when applied appropriately} 
\tool improves the limitations of previous black-box generative-based test generators~\cite{weissl2024targeted,maryam2025benchmarking} by employing a white-box approach that facilitates controllability without retraining. Particularly, \tool avoids direct noise-space perturbations and instead operates through semantically grounded transformations that remain within the model's operating regime. On the other hand, \tool adapts the ControlNet architecture and inverts its control mechanism: rather than supplying a fixed control signal, \tool iteratively modifies the generated image until the observed control signal enters a targeted failure region.
This design enables \tool to generate realistic, high-quality test cases without the overhead of constructing datasets of failures. As a result, \tool achieves runtime efficiency comparable to prior approaches, even when contrasted with the smaller models used in \mimicry (\autoref{tab:runt}), while improving test-case realism (\autoref{tab:rq3_im}) and overall effectiveness in generating failures for specified inputs (\autoref{tab:merged_rq1}).

\head{\tool is useful for various testing domains with varying complexity of SUT behavior} As AI becomes more important in diverse systems, its functionality can vary drastically depending on the task; while previous work has mostly focused on classification, other tasks such as object detection or binary classification are equally relevant. In our work, we show that \tool excels in testing different SUTs with fewer restrictions than our baselines. In comparison to \mimicry, \tool does not rely on the presence of a conditional generative network, as it is gradient based and can therefore change attributes in images that are not explicitly ''targetable'' by the generative network itself. This allows general-purpose networks such as pretrained Stable Diffusion models to be used even when they do not allow specific targeting toward semantic conditions. As shown in the \celeba task, this reliance on explicit targeting makes \mimicry fail, as without a target of the required semantic class the optimization has no hope of succeeding (\autoref{tab:merged_rq1}). Similarly, in \mrm, the conditioning is enforced through specific prompts, which limits testing to domains with available datasets; for example, in \celeba one would need a dataset describing facial attributes in each image with great detail to generate an image containing the desired attribute. In contrast, \tool can generate any image using the same \textit{unconditioned} generator and manipulate it such that the attribute appears or disappears based on its oracle, the SUT. Conversely, we also explore testing of object detectors such as \yolo in complex domains like self-driving, where semantically rich scenarios are difficult to specify and produce datasets for; here \tool excels in its ability to perform targeted manipulations, whereas baselines fail and resort to global corruption (\autoref{fig:yolo_examples}).

\head{On practical strategies for selecting \tool learning-rate schedules}
Although the optimal learning-rate schedule differs across datasets and SUTs, the trade-off (\autoref{fig:im_sweep}\& \autoref{fig:sweep_corr}) curves presented in \autoref{sec:sens} provide a practical heuristic for selecting suitable learning schedules for new domains. In particular, the sensitivity analysis consistently shows that excessively small learning rates substantially increase runtime while providing only marginal improvements in realism and structural preservation. Conversely, overly large learning rates reduce runtime but introduce increasingly unrealistic perturbations. Across all evaluated domains, appropriate schedules can therefore be identified by selecting the transition region in which runtime stabilizes while image realism and structural similarity remain largely preserved. Furthermore, the strong correlation between human realism ratings and FID enables automated identification of these trade-off regions without requiring extensive manual inspection.

%% file: sec/6_related_work.tex
\section{Related Work}\label{sec:related}

Deep learning test generation techniques fall broadly into three families: model-based input manipulation, raw input manipulation, and latent-space manipulation. We briefly outline each to position \tool within the state of the art. Our emphasis is on approaches that can generate high-fidelity, in-distribution test cases for complex systems—an area where existing methods remain limited.

Model-based input manipulation (MIM) constrains test generation to a manually crafted representation of the input domain~\cite{Gambi:2019:ATS:3293882.3330566,riccio2020deepjanus,isa, zohdinasab2021deephyperion, 2021-Riccio-ASE,2026-Chen-EMSE}. These techniques can produce valid inputs when such domain models are available but are inherently restricted to simple or well-structured domains where clear input models exist. As demonstrated in \mimicry ~\cite{weissl2024targeted}, MIM approaches struggle to scale to more complex, high-dimensional data, and typically fail to generate realistic test cases. Their reliance on hand-engineered input abstractions limits applicability in realistic settings where the data distribution cannot be explicitly modeled.

Raw input manipulation (RIM) operates directly in pixel space by perturbing existing images, either by exploiting knowledge of internal parameters of the SUT in a white-box setting~\cite{im2022adversarial,kurakin2018adversarial,croce2020minimally}, or by applying optimization-based techniques in a black-box manner~\cite{liu2022deepboundary,zhang2020deepsearch}. While effective for evaluating adversarial robustness, RIM methods typically do not generate functionally novel test cases. Instead, they rely on small, often imperceptible perturbations to induce model failures, producing samples that may lie outside the true data distribution or exhibit adversarial artifacts~\cite{pei2017deepxplore,guo2018dlfuzz,deeptest}. As a result, RIM is well suited for robustness assessment but less appropriate for functional test generation, where the objective is to synthesize realistic, in-distribution inputs that reflect semantically meaningful variations and expose model-level functional failures~\cite{buzhinsky2023metrics}. Gradient-based approaches are also highly effective and computationally efficient for adversarial robustness evaluation, particularly in white-box settings where direct gradient access enables targeted failure induction~\cite{buzhinsky2023metrics}. However, because these methods primarily optimize prediction sensitivity through low-level perturbations, they are less suited for evaluating the generalization ability of SUTs under semantically meaningful distribution shifts, which is the primary focus of this work~\cite{maryam2025benchmarking, maryam2026deepnaqqal}.

Latent–space manipulation methods have gained traction as alternatives to MIM and RIM, whose effectiveness diminishes in complex, information-rich domains~\cite{2026-Merabishvili-ICSEW,2026-Maryam-SCICO}. These approaches exploit the latent spaces of generative models, which implicitly capture the structure of the underlying data distribution with far greater expressiveness than manually engineered models.

Several methods perturb latent representations using noise. Sinvad applies noise-based perturbations in the latent space of VAEs~\cite{kang2020sinvad}, though VAEs generally lack the expressiveness needed for complex data domains~\cite{weissl2024targeted}. Buzhinsky et al.~\cite{buzhinsky2023metrics} extend this idea to GANs, benefiting from higher-quality generation but still offering limited control. More recent work such as \mrm~\cite{maryam2025benchmarking} generalizes noise-based manipulation across VAEs, GANs, and diffusion models by injecting noise into intermediate latent representations. While \mrm can operate on all three model classes, we employ only its diffusion-based variant because diffusion latents provide substantially higher fidelity and stability than the VAE and GAN counterparts. Even in diffusion space, however, \mrm remains an untargeted perturbation method that offers limited fine-grained control over generated test cases.

\mimicry~\cite{weissl2024targeted}, one of our baselines, pursues a different latent-space strategy. It samples origin and target seeds in StyleGANs latent $w$-space, interpolates between them, and optimizes the interpolation weights to steer generation toward targeted test cases such as boundary or adversarial samples. This yields more directed behavior than noise-based methods while remaining fully dataset-free. Similarly, \textsc{Detect}~\cite{2026-Chen-DETECT} works with StyleGANs, but in the feature space, instead of the whole input space, which allows more fine-grained feature-based perturbations. 
% Although proposed for elicitation of spurious features it has been compared to boundary testing cases which in which it outperforms \mimicry.

Bao et al.~\cite{bao2024generative} take a broader approach by incorporating generative models directly into the SUT's training loop. Using VAEs, GANs, and diffusion models, they generate previously unseen samples that are then treated as test cases without specific conditioning.

Diffusion-based methods have become particularly prominent due to their strong generative performance~\cite{maryam2025benchmarking, missaoui2023semantic, bao2024generative}. Missaoui et al.~\cite{missaoui2023semantic} use diffusion models for semantic control via targeted inpainting to synthesize failure-inducing inputs. Yet explicit conditioning and fine-grained behavioral control remain limited: existing techniques rely on untargeted perturbations or surrogate models that constrain manipulations to semantically valid regions.

To address these limitations, we propose \tool, a dataset-free diffusion-based method enabling direct, behavior-level control over generated test cases without requiring a surrogate model for inpainting. 
% Although our focus is on image-based methods, latent-space manipulation has also been extended to other modalities, including text generation~\cite{ren2020generating}.

%% file: sec/7_conclusion.tex
\section{Conclusion}

In this work we present \tool, a diffusion-based test-case generator that uses HyperNetwork adaptation. \tool enables dataset-free and controllable test-case generation for a variety of deep learning systems, by adapting the ControlNet architecture and a customized tuning regime. It achieves higher performance in both test discovery and test-case realism compared to relevant baselines. We compare \tool to \mimicry, a state-of-the-art test-case generation method based on StyleGANs, and to \mrm, a complementary diffusion-based approach. While these baselines struggle with generation realism and fine-grained control, \tool performs reliably across all tasks by respecting the model's internal structure and operational constraints.

Our empirical study shows that \tool has similar runtime on our hardware but significantly outperforms the baselines in terms of both the quantity of discovered test cases and their quality. Human evaluators rate \tool's test cases as more realistic and more semantically and structurally meaningful.

We argue that test-case generation for deep learning models is only useful when the generated cases cover a broad range of increasingly complex scenarios. Simple examples (e.g., digit classification) are insufficient for evaluating modern systems. Future work should therefore target visually complex domains such as automated driving, or evaluate systems under test that are inherently more robust to small perturbations, such as object detectors or segmentation models. Another promising direction is refining the balancing of HyperNet, which becomes increasingly important as data complexity grows and strongly influences output quality. Future work could additionally investigate whether the generated failure-inducing test cases can be leveraged for robustness retraining of SUTs after appropriate semantic validation of the generated samples.

\section{Data Availability}\label{sec:da}
The codebase, analysis scripts, and all artifacts generated for \tool and \mimicry experiments are available in the replication package~\cite{replication-package}. For replication of \mrm experiments, please refer to the original publication~\cite{maryam2025benchmarking}.

%% file: sec/8_appendix.tex
\appendix
\raggedbottom
\section{Methodology}
\subsection{Derivation of the $\ell_1$ Formulation of Trace Difference}
\label{app:l1trace}
We show that the trace-based definition of $\operatorname{trace\Delta}$ in \autoref{eq:trace} is equivalent to an $\ell_1$-distance computed on the diagonal of the covariance difference. Let
\[
\Sigma_0 = \operatorname{cov}(E_0^\intercal),
\qquad
\Sigma_t = \operatorname{cov}(E_t^\intercal),
\]
with $\Sigma_0, \Sigma_t \in \mathbb{R}^{d \times d}$.
By definition, the trace of a matrix is the sum of its diagonal entries:
\[
\operatorname{tr}(A) = \sum_{i=1}^d A_{ii}.
\]
Applying the trace to the elementwise absolute difference of the covariance matrices $\Sigma_\Delta = \Sigma_0 - \Sigma_t$ gives
\[
\operatorname{tr}\!\left(\left|\Sigma_\Delta\right|\right)
=
\sum_{i=1}^d \left|(\Sigma_\Delta)_{ii}\right|.
\]
Equivalently, isolating the diagonal via a Hadamard product with the identity matrix $I$ yields the diagonal matrix 
\[
D := \Sigma_\Delta \odot I,
\qquad\text{so that}\qquad
D_{ij} = (\Sigma_\Delta)_{ij}\, I_{ij}
=
\begin{cases}
(\Sigma_\Delta)_{ii} & i=j,\\
0 & i\neq j.
\end{cases}
\]
Using the entry-wise $\ell_1$-norm $\|A\|_{1} := \sum_{i,j}|A_{ij}|$, we obtain
\[\| \Sigma_\Delta \odot I \|_{1}
= \sum_{i=1}^d\sum_{j=1}^d |D_{ij}|
= \sum_{i=1}^d |D_{ii}|
= \sum_{i=1}^d |(\Sigma_\Delta)_{ii}|
= \operatorname{tr}\!\left(\left|\Sigma_\Delta\right|\right).\]
Thus, $\operatorname{trace\Delta}$ equals the $\ell_1$-distance between the diagonals of the covariance matrices, capturing per-dimension changes while discarding cross-dimensional correlations.

\section{Study}
\begin{figure}[H]
    \centering
    \begin{subfigure}[b]{0.32\columnwidth}
        \centering
        \includegraphics[width=\linewidth]{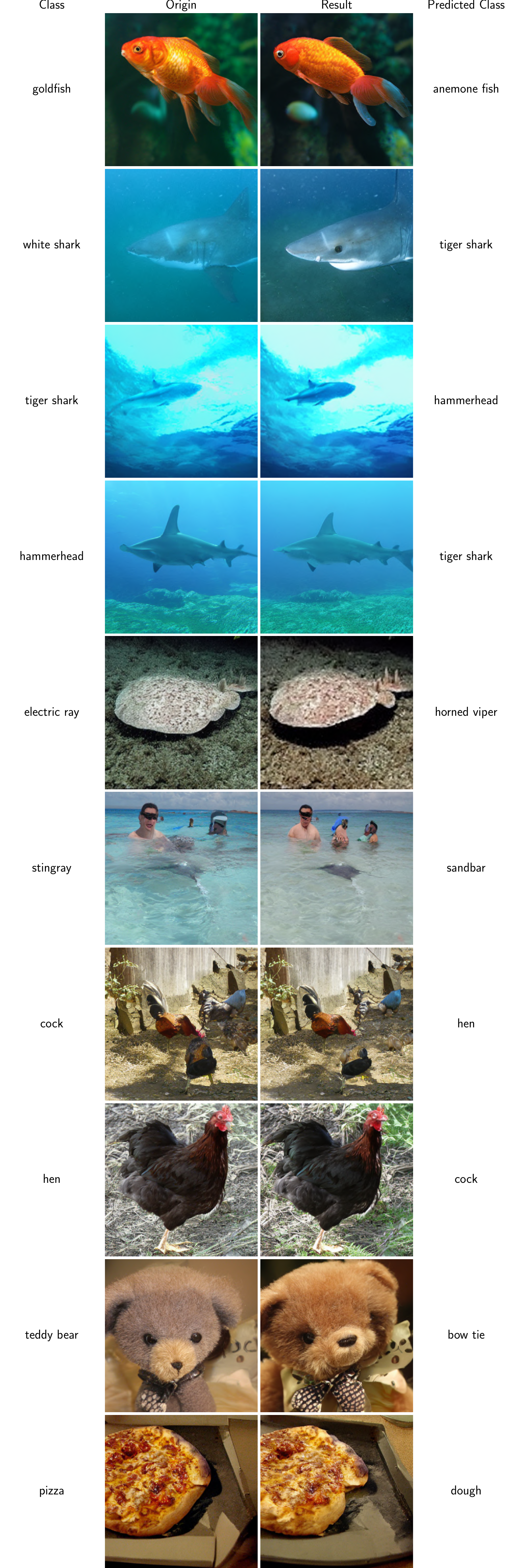}
        \caption{\tool}
        \label{fig:ex_hyn_i}
    \end{subfigure}
    \hfill
    \begin{subfigure}[b]{0.32\columnwidth}
        \centering
        \includegraphics[width=\linewidth]{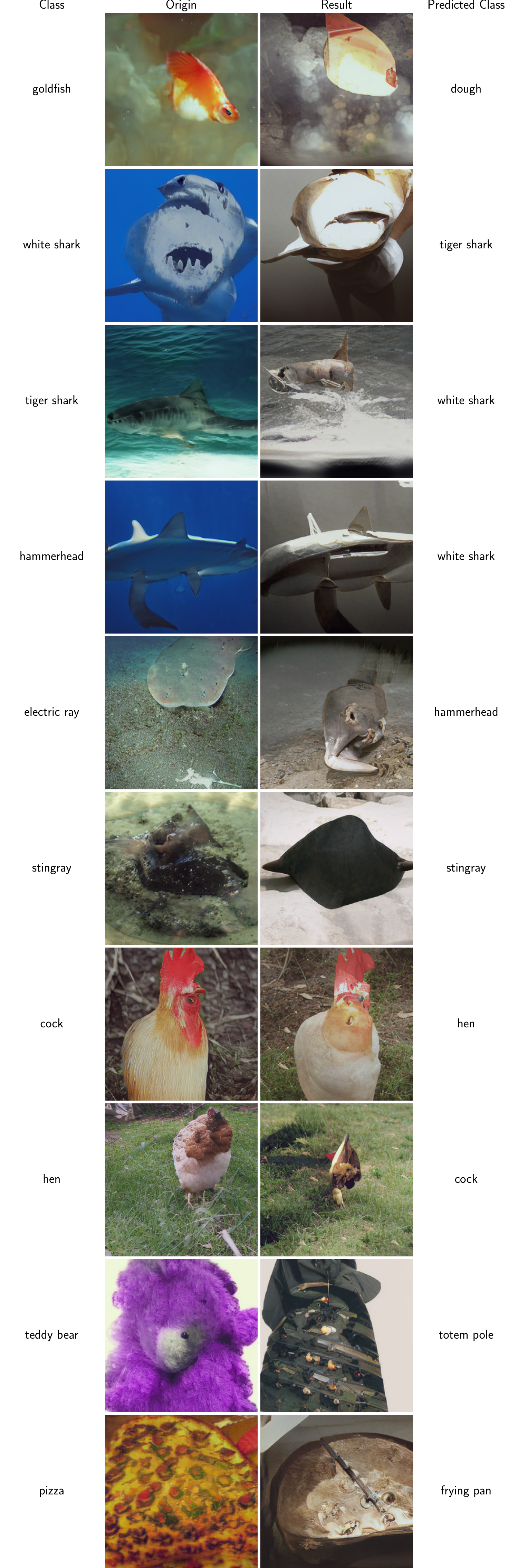}
        \caption{\mimicry}
        \label{fig:ex_mim_i}
    \end{subfigure}
    \hfill
    \begin{subfigure}[b]{0.32\columnwidth}
        \centering
        \includegraphics[width=\linewidth]{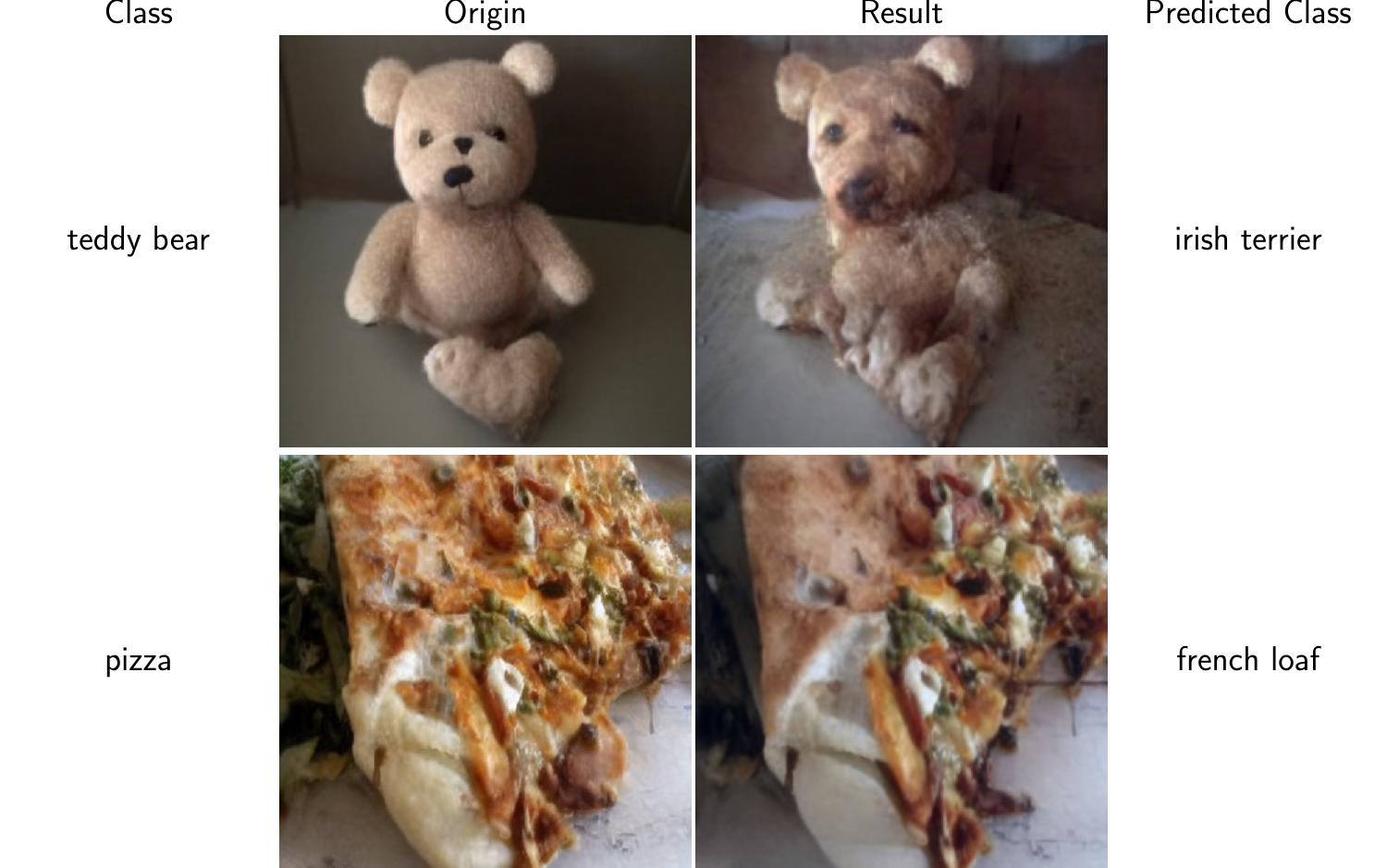}
        \caption{\mrm}
        \label{fig:ex_mrm_i}
    \end{subfigure}

    \caption{Example Origins and Targets for \imagenet[icon]. Predicted Origin and Target class are shown to the left and right of the images.}
    \label{fig:imagenet_examples}
\end{figure}

\begin{figure}[H]
    \centering
    \begin{subfigure}[b]{0.32\columnwidth}
        \centering
        \includegraphics[width=\linewidth]{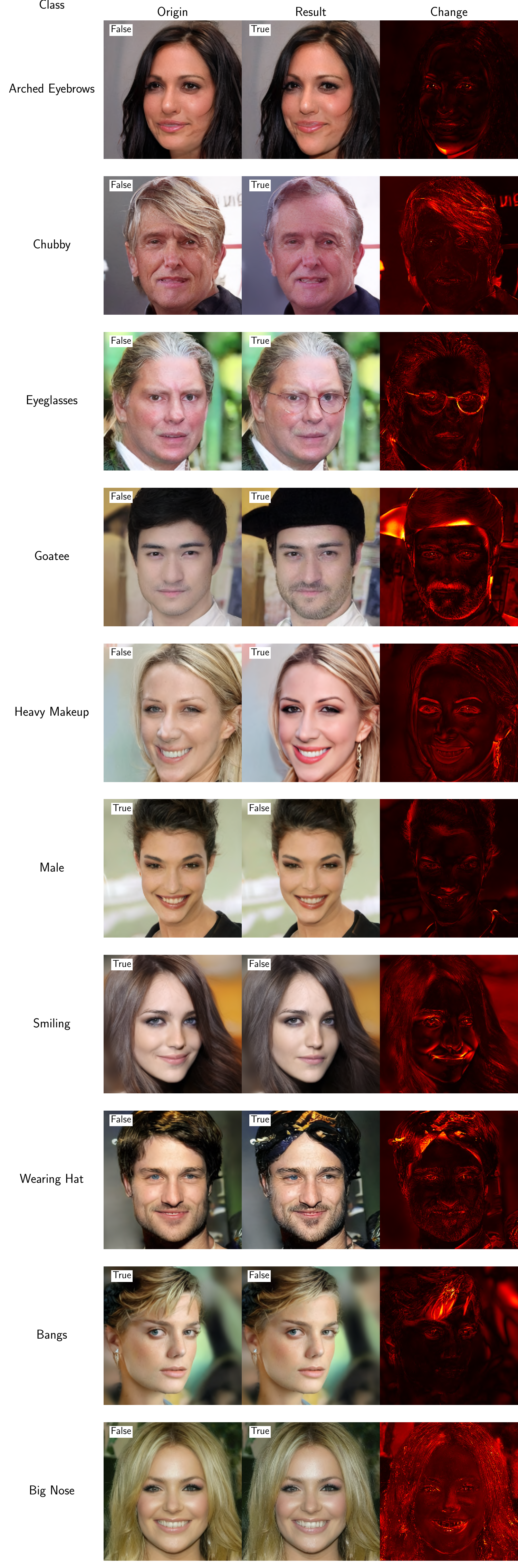}
        \caption{\tool}
        \label{fig:ex_hyn_c}
    \end{subfigure}
    \begin{subfigure}[b]{0.32\columnwidth}
        \centering
        \includegraphics[width=\linewidth]{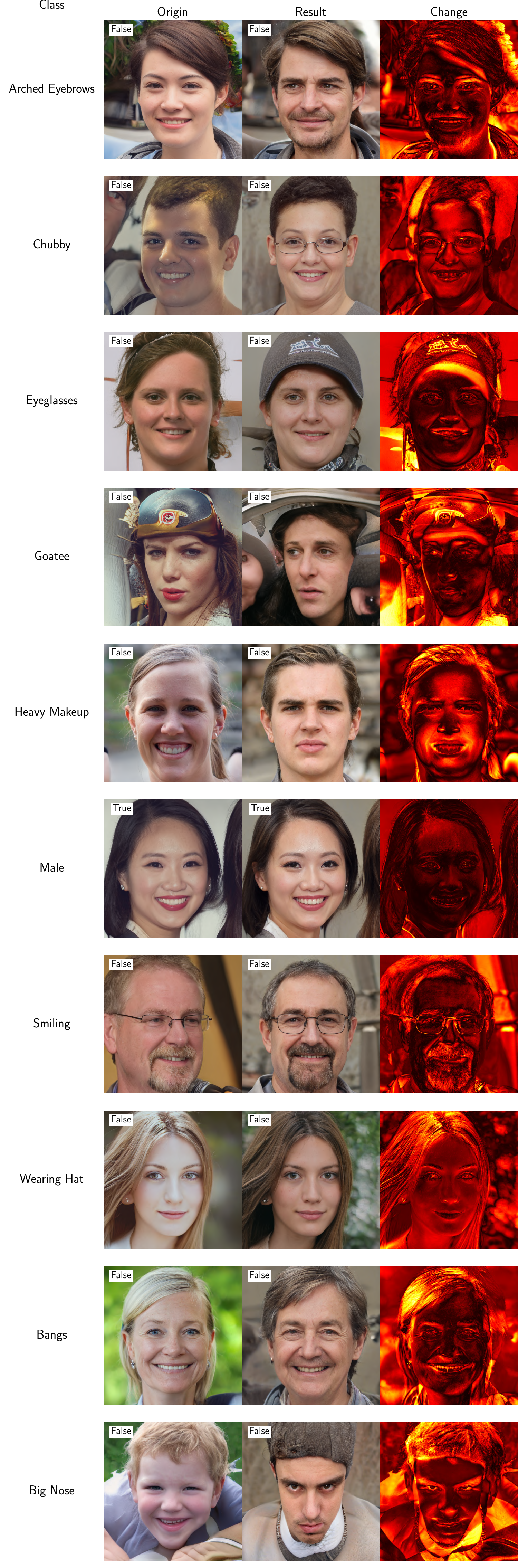}
        \caption{\mimicry}
        \label{fig:ey_mim_c}
    \end{subfigure}
    \begin{subfigure}[b]{0.32\columnwidth}
        \centering
        \includegraphics[width=\linewidth]{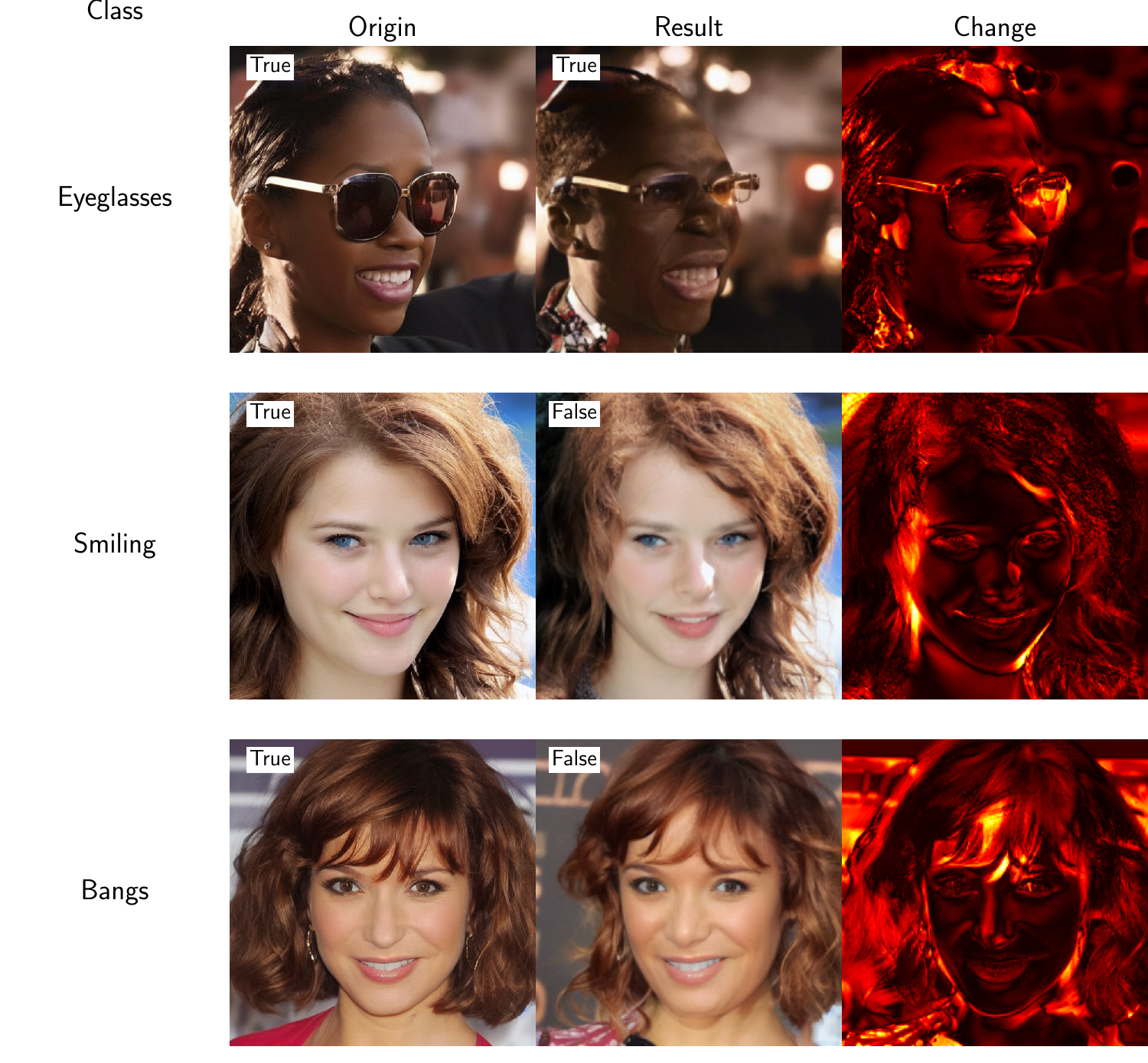}
        \caption{\mrm}
        \label{fig:ey_mrm_c}
    \end{subfigure}
    \caption{Example Origins and Targets for \celeba[icon]. Predicted attribute presence is shown in the top left corner of the examples, the difference between Origin and Target is shown in the last column.}
    \label{fig:celeba_examples}
\end{figure}

\begin{figure}[H]
    \centering
    \begin{subfigure}[b]{0.47\columnwidth}
        \centering
        \includegraphics[width=\linewidth]{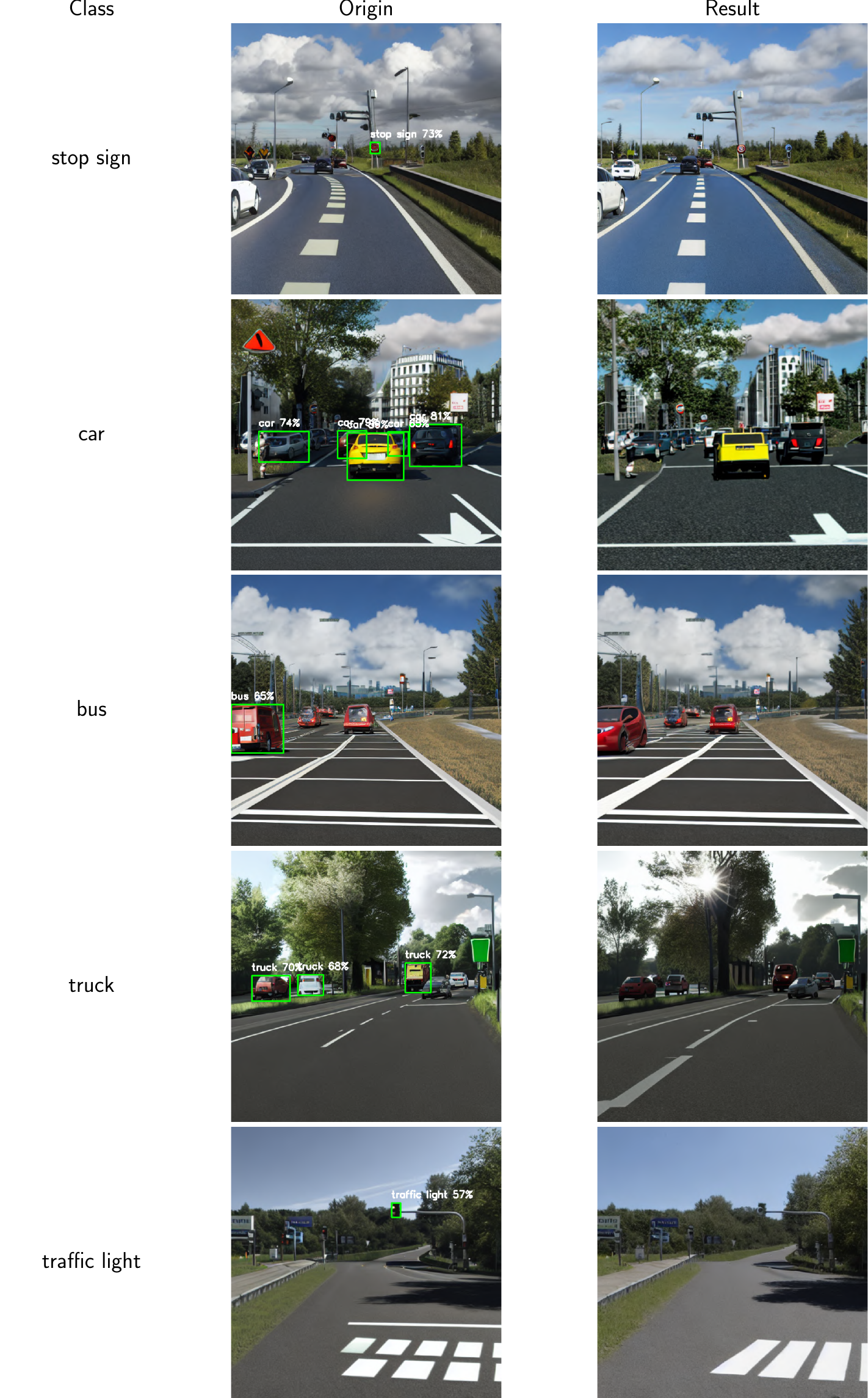}
        \caption{\tool}
        \label{fig:ex_hyn_y}
    \end{subfigure}
    \begin{subfigure}[b]{0.47\columnwidth}
        \centering
        \includegraphics[width=\linewidth]{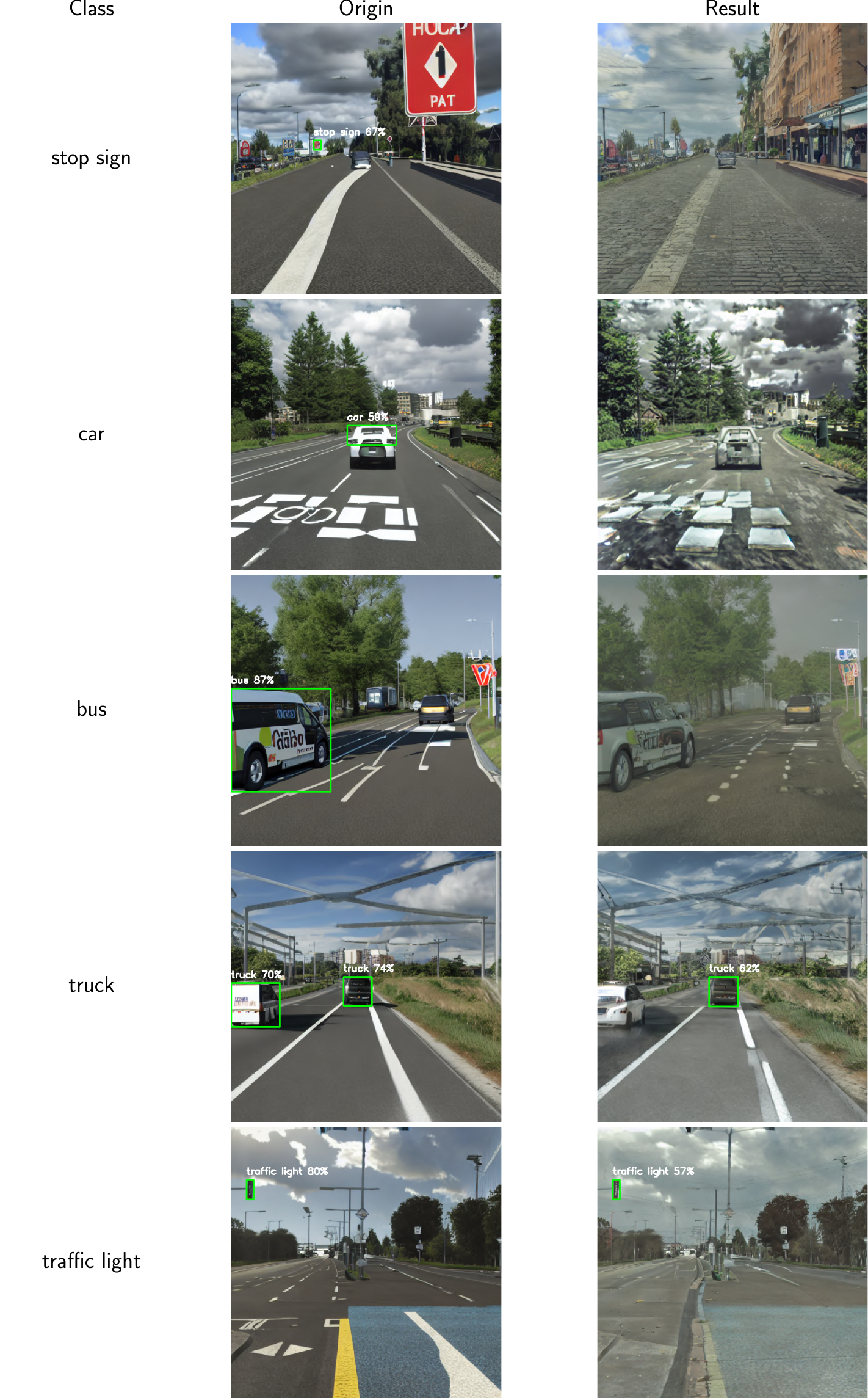}
        \caption{\mrm}
        \label{fig:ex_mrm_y}
    \end{subfigure}
    \caption{Example Origins and Targets for the \yolo[icon] Dataset. Predicted Objects are shown with bounding boxes and their confidence. NMS was used to reduce duplicate bounding boxes and a confidence threshold of 0.5 is used to filter further.}
    \label{fig:yolo_examples}
\end{figure}

\begin{promptbox}\label{prompt:posp}
\textit{"A photorealistic urban traffic scene with cars, traffic lights, and stop signs, clear skies, daytime, featuring a \{class\}"}    
\end{promptbox}

\begin{promptbox}\label{prompt:negp}
\textit{"blurry, distorted, ugly, low quality, cartoon, sketch"}
\end{promptbox}
\newpage
\section{Potential Pitfalls of Noise-Based Latent Manipulation in LDMs}\label{sec:distrshift}
In LDMs, the latent code to denoise $z\in\mathbb R^d$ is assumed to follow a standard Gaussian prior,
\[
z \sim \mathcal N(0, I_d),
\]
so that the decoder (usually a VAE) reliably reconstructs codes consistent with that prior.  

In the perturbation scheme used in \mrm, latents are iteratively mutated via
\[
z_{k+1} = z_k + \epsilon_k, \quad \epsilon_k \sim \mathcal N(0, \delta_k I_d),
\]
where initially
\[
\delta_k = \alpha R, \quad
\alpha = \begin{cases}
10^{-3} & \text{if \textit{config: low}}\\
10^{-4} & \text{if \textit{config: high}}
\end{cases},
\]
with $R$ derived from the observed range of initial latents. $\delta_{k+1}$ may double if fitness does not improve in a generation, or reset to $\delta_0$ if it does.  

Over multiple iterations, the latent distribution spreads due to additive noise:
\[
z_k \sim \mathcal{N}\Big(0,\, I_d + \sum_{i=1}^k \delta_i I_d\Big).
\]
But since \mrm uses clipping to keep perturbed latents in the initial range $z_{\min} \le z_k \le z_{\max}$, the resulting distribution is no longer Gaussian, and higher-order moments (kurtosis) deviate from the original prior. Consequently, many latents still fall outside the region where the decoder was trained to operate reliably, making them effectively out-of-distribution (OOD). 

\begin{figure}[H]
    \centering
    \includegraphics[width=0.85\linewidth]{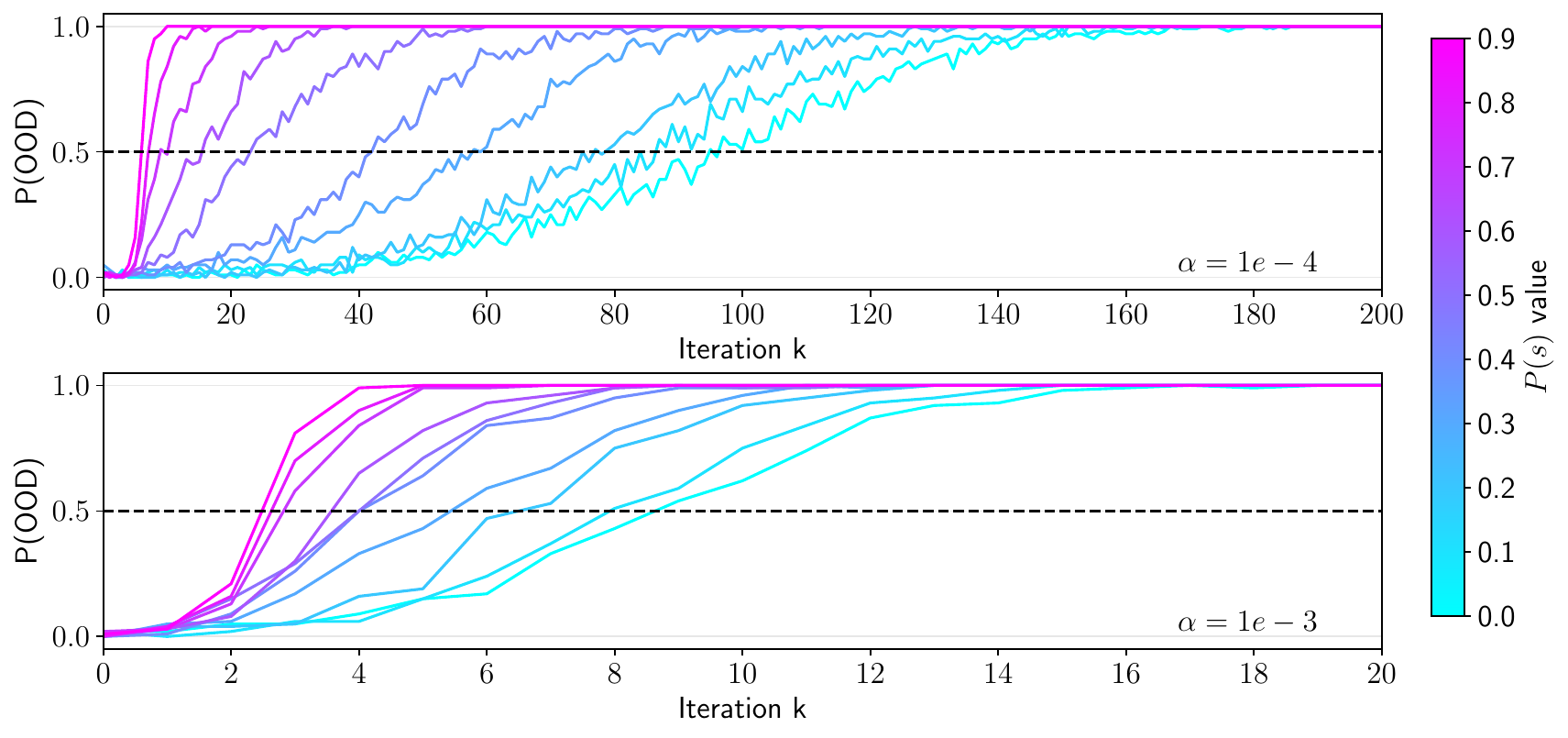}
    \caption{Probability of OOD latents over iterations in \mrm for both \textit{low} (top) and \textit{high} (bottom) configurations. Latents are considered OOD if their squared norm exceeds the 99th percentile of the $\chi^2(d)$ distribution. $P(s)$ is the probability of scaling $\alpha$, where higher values mimic a weaker SUT.}
    \label{fig:appx}
\end{figure}

As shown in \autoref{fig:appx}, even when scaling $\alpha$ is unnecessary, latents eventually exhibit a high probability of being OOD, making the generator unreliable. Unlike classical genetic algorithms, where fitness penalizes undesirable behavior, here decoder failure is not penalized. Consequently, the optimization process can exploit these "bad" latents, producing examples that appear successful according to the optimization objective but are semantically invalid.  

For our experiments, the number of iterations $k$ averaged 28 for \imagenet and 9 for \yolo (\autoref{tab:runt}, 25-budget per iteration), both resulting in a high probability ($>0.5$) of OOD when using the perturbation sizes provided in \mrm's GitHub with \textit{config: high}\cite{maryam2025benchmarking}.

\begin{figure}[H]
    \centering
    \includegraphics[width=0.85\linewidth]{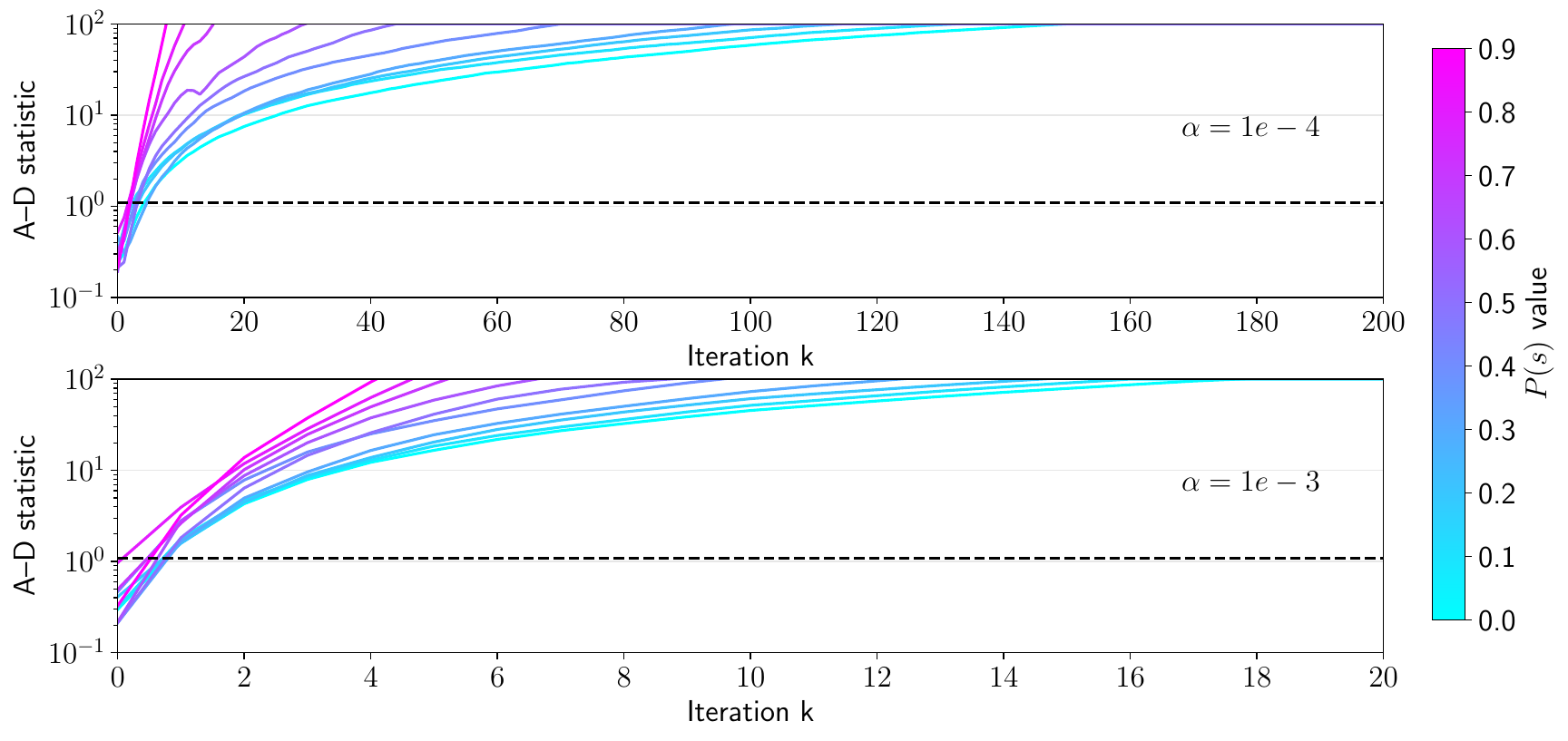}
    \caption{The Anderson–Darling test statistic with a a criticality threshold of $1.092$ ($p < 0.01$) indicated by the dotted black line. Experiments across multiple scaling probabilities $P(s)$ are shown.}
    \label{fig:scaling}
\end{figure}

Looking at \autoref{fig:scaling}, we can see that clipping highly variable Gaussian noise disrupts the normality of the data, as confirmed by the Anderson–Darling test, which properly emphasizes the tails. A higher test statistic indicates a lower probability of the data being normally distributed; in our case, the significance threshold is exceeded for most configurations within the first $5\%$ of iterations. Previous works have shown that the type of noise has a substantial impact on generation quality~\cite{nachmani2021non, vandersanden2024edge, shariatianheavy}, with both the symmetry of the noise and its inversion being directly correlated with generation quality~\cite{qi2024not}. Clipping partially breaks this symmetry for inversion, providing a likely explanation for the observed degradation in output quality (\autoref{fig:ex_mrm_i} \& \autoref{fig:ex_mrm_y} \& \autoref{fig:ey_mrm_c}).